\documentclass[9.5pt,journal,final]{IEEEtran}

\usepackage{graphicx}
\usepackage{epsfig}
\usepackage{amsmath}
\usepackage{mathtools}
\usepackage{xspace}
\usepackage{algorithm}
\usepackage{amssymb}
\usepackage{bm}
\usepackage{bbding}
\usepackage{booktabs} % \midrule
\usepackage{tabularx}
\usepackage{multirow}
\usepackage{multicol}
\usepackage{etoolbox,siunitx}
\usepackage{color}      % for illustrations
\usepackage[dvipsnames]{xcolor}
\usepackage{diagbox}
\usepackage[font=scriptsize,skip=2pt,subrefformat=parens]{subcaption}
\usepackage[colorlinks=true, allcolors=blue]{hyperref}
\usepackage{ragged2e}
\usepackage{float}
\usepackage{bbding}
\usepackage{amsfonts}
\usepackage{utfsym}

\usepackage{algpseudocode}
\usepackage{listings}
\usepackage{wrapfig}
\usepackage{makecell}

\usepackage[utf8]{inputenc}
\ifCLASSINFOpdf
\fi

\usepackage[utf8]{inputenc}

\lstdefinestyle{Python}{
  language=Python,
  basicstyle=\ttfamily\small,
  commentstyle=\color{YellowGreen},
  keywordstyle=\color{blue}\bfseries,
  stringstyle=\color{red},
  showstringspaces=false,
  breaklines=true,
  breakatwhitespace=true,
  frame=single,
}

\newcommand{\ie}{\textit{i}.\textit{e}.}
\newcommand{\eg}{\textit{e}.\textit{g}.}

\begin{document}
\title{Cross-level Multi-instance Distillation for Self-supervised Fine-grained Visual Categorization}
%\title{Learning Local Semantics for \\Aerial Scene Recognition}

\author{Qi~Bi,~\IEEEmembership{Member,~IEEE,}
        Wei~Ji$^*$,
        Jingjun~Yi,
        Haolan~Zhan,
        Gui-Song~Xia$^*$,~\IEEEmembership{Senior~Member,~IEEE}% <-this % stops a space

\thanks{Qi Bi is now with University of Amsterdam, 1098XH, the Netherlands. He was with Wuhan University, Wuhan, 430072, China.}% <-this % stops a space
\thanks{Jingjun Yi is with Wuhan University, Wuhan, 430072, China.}% <-this % stops a space
\thanks{Wei Ji is with Yale University, New Haven, the United States.}
\thanks{Haolan Zhan is with Monash University, Australia. }
\thanks{G.-S. Xia is with the School of Artificial Intelligence, National Engineering Research Center for Multimedia Software, School of Computer Science, and also the State Key Lab. LIESMARS, Wuhan University, Wuhan, 430072 China. $^*$Correspondence: {\em wei.ji@yale.edu} and {\em guisong.xia@whu.edu.cn}.}
% <-this % stops a space
}
\maketitle

\begin{abstract}
High-quality annotation of fine-grained visual categories demands great expert knowledge, which is taxing and time consuming.
Alternatively, learning fine-grained visual representation from enormous unlabeled images (\eg, species, brands) by self-supervised learning becomes a feasible solution.
However, recent investigations find that existing self-supervised learning methods are less qualified to represent fine-grained categories.
The bottleneck lies in that the pre-trained class-agnostic representation is built from every patch-wise embedding, while fine-grained categories are only determined by several key patches of an image.
In this paper, we propose a \textbf{C}ross-level \textbf{M}ulti-instance \textbf{D}istillation (CMD) framework to tackle this challenge.
Our key idea is to consider the importance of each image patch in determining the fine-grained representation by multiple instance learning.
To comprehensively learn the relation between informative patches and fine-grained semantics,
the multi-instance knowledge distillation is implemented on both the region/image crop pairs from the teacher and student net, and the region-image crops inside the teacher / student net, which we term as \textit{intra-level multi-instance distillation} and \textit{inter-level multi-instance distillation}. 
Extensive experiments on several commonly used datasets, including CUB-200-2011, Stanford Cars and FGVC Aircraft, demonstrate that the proposed method outperforms the contemporary methods by up to 10.14\% and existing state-of-the-art self-supervised learning approaches by up to 19.78\% on both top-1 accuracy and Rank-1 retrieval metric. 
Source code is available at \url{https://github.com/BiQiWHU/CMD}.%Source code will be publicly available.
\end{abstract}

\begin{IEEEkeywords}
Fine-grained Visual Categorization, Self-supervised Learning, Multiple Instance Learning, Knowledge Distillation.
\end{IEEEkeywords}

% \fi
\IEEEpeerreviewmaketitle

\section{Introduction}
\label{S1}

\subsection{Problem Statement}

\IEEEPARstart{F}{ine}-grained visual categorization (FGVC) aims to discern similar objects, brand and species from the same coarse-grained category to their individual fine-grained categories \cite{han2022pcnn,wang2022knowledge,luo2019cross,ge2019weakly,Song2023SEB}. 
A key bottleneck of FGVC is the scarcity of large-scale high-quality fine-grained annotation, which costs extensive time and effort for experts with proficient domain knowledge \cite{Wei2022survey}.
While the amount of sample annotation is large-scale in many vision tasks \cite{bi2023interactive,pan2022label}, the sample amount of most existing FGVC datasets is still under ten-thousand level.
As there are huge amount of unlabeled images from similar species, brands and objects, self-supervised learning (SSL) techniques, which learn the class-agnostic representation from unlabeled images \cite{chen2020simple,chen2021exploring,chen2020improved,caron2021emerging}, provide a feasible substitution to alleviate this dilemma.

\begin{figure}[!t]
    \centering
    \includegraphics[width=1.0\linewidth]{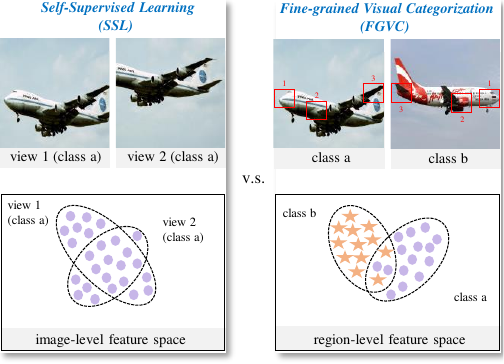}
    \vspace{-0.4cm}
    \caption{Key challenge to incorporate self-supervised learning (SSL) for fine-grained visual categorization (FGVC) lies in the learning objective inconsistency. \textbf{left}: SSL leverages image-level objectives to learn the same coarse-grained feature space; \textbf{right}: FGVC leverages region-level objectives to learn subtle parts (in \textcolor{red}{red boxes}) to discern an image from other fine-grained categories under the same coarse-grained category.}
    %\vspace{-0.3cm}
    \label{challenge}
\end{figure}

Recent works have explored the possibility, but unfortunately found that existing SSL pipelines are less qualified to highlight the key part patterns (\eg, informative patches) for fine-grained semantics \cite{cole2022does,shu2023learning}. 
Specifically, an indiscriminative FGVC feature space from contrastive learning is observed \cite{cole2022does}, and a number of dominant SSL pipelines show significantly inferior performance compared with the fully-supervised upper bound \cite{shu2023learning}.
Overall, adapting SSL for FGVC remains intriguing but less explored.

The key challenge of \textit{self-supervised fine-grained visual representation learning} lies in the learning objective inconsistency between FGVC and SSL.
Existing SSL pipelines leverage image-level learning objectives (illustrated in Fig.~\ref{challenge}~left), which minimize the distances between two augmented views by taking the feature embedding from an image into account \cite{cole2022does}.
In contrast, FGVC relies on region-level learning objectives (illustrated in Fig.~\ref{challenge}~right).
As the fine-grained categories from a certain coarse-grained category only have subtle differences, the fine-grained semantics are usually only determined by some key local regions. 
Thus, the information from the key part patterns can be overwhelmed from the huge amount of indiscriminative patterns by existing SSL pipelines.

\begin{figure}[!t]
    \centering
    \includegraphics[width=1.0\linewidth]{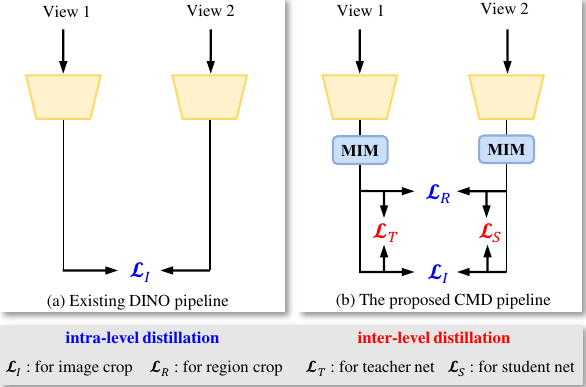}
    \vspace{-0.4cm}
    \caption{Flowchart of the existing DINO paradigm (left) and the proposed \textbf{C}ross-level \textbf{M}ulti-instance \textbf{D}istillation CMD (right). MIM refers to multi-instance modeling. }
    %\vspace{-0.3cm}
    \label{motivation}
\end{figure}

\subsection{Motivation \& Objectives}

In general, the pioneering self-supervised fine-grained visual categorization methods either directly apply the image-level contrastive learning objective \cite{cole2022does} or incorporate the region-level activation patterns when doing inference on the acquired contrastive representation \cite{shu2023learning}.
Unfortunately, the aforementioned learning objective inconsistency is not essentially addressed.
In this paper, we are motivated to push the frontier of \textit{self-supervised fine-grained visual representation learning}
by addressing the learning objective inconsistency between SSL and FGVC.
In contrast to existing methods \cite{cole2022does,shu2023learning}, our learning objective is devised in an innovative way such that, the class-agnostic representation can not only focus on the discriminative regions when learning, but also explore the relation between the region patterns and the image-level context.
The specific objectives are two-fold.

\subsubsection{The key local regions that determine the fine-grained categories contribute more to the pre-trained class-agnostic representation}
This objective is intuitive and straight-forward to understand. 
Only when the subtle fine-grained patterns, which rest in several local regions of an image, are highlighted in the SSL pipeline, can the pre-trained class-agnostic representation carries more discriminative information for fine-grained patterns.

To realize this objective, multiple instance learning (MIL) is introduced \cite{wei2019multi,ilse2018attention,wang2018revisiting} to model the relation between image and each patch, where each patch is regarded as an instance and each image is regarded as a bag \cite{wang2013max,wang2015relaxed,tang2017multiple}. 
In contrast to the conventional MIL with available bag-level annotation \cite{ilse2018attention,wang2018revisiting}, the bag annotation is unavailable in our self-supervised fine-grained learning problem. 
To this end, the bag representations from the augmented crop pairs are learned under the self-supervised knowledge distillation without label (DINO) paradigm \cite{caron2021emerging}, so as to guide the instance-level representation learning. 

\subsubsection{Aligning the fine-grained information between region-level and image-level}
Modern SSL pipelines (\eg, EsViT \cite{li2021efficient}) use both region- and image- level crops to learn pre-trained class-agnostic representation, which provides more access to extensively exploit the informative patches corresponding to the fine-grained semantics. 
Ideally, the cropped image-region
pairs that represent similar fine-grained categories need to be close together, while those represent
dissimilar fine-grained categories need to be far apart.

To realize this objective, we propose both intra-level and inter-level multi-instance distillation (in Fig.~\ref{motivation}). 
The intra-level multi-distillation acquires the fine-grained knowledge from image-level crops and region-level crops from the teacher-student pair. 
In contrast, the inter-level multi-instance distillation acquires the fine-grained knowledge from image-region crop pairs inside the teacher/student net.
The joint distillation of both helps comprehensively learn the relation between informative parts and fine-grained semantics of an image.

\subsection{Contribution}

Compared with existing FGVC methods that leverage intra-/inter- level relation, the propose method makes the following advancements.
Firstly, while existing methods rely on the fine-grained semantic annotation to constrain the intra-/inter- relation, the proposed method innovatively constrains the intra-/inter- level representation by the same fine-grained semantics without any label supervision.
Secondly, existing methods do not consider the aforementioned learning objective inconsistency problem in self-supervised FGVC. 
In contrast, the proposed method resolves it by incorporating the regions of fine-grained patterns to the self-supervised representation, and constrains both regions and images under the same fine-grained semantics.
Thirdly, the proposed method leverages the multiple instance learning (MIL) formulation to theoretically realize this constraint. MIL has been rarely used for self-supervised learning and self-supervised fine-grained representation, where the proposed method bridges the research gap. 

From a technical perspective, existing knowledge distillation with no label (DINO) paradigm only learns a global representation from image-level crops.
In contrast, the proposed method advances DINO by both image-level and region-level crops, so as to perceive the subtle fine-grained patterns from the region while at the same time be constrained by the context information from the image.
Besides, it makes the first exploration to  simultaneously model the relation for both region- and image- level bags by MIL formulation.
Further, the proposed intra- and inter- level multi-instance distillation, under the strict MIL formulation, allow both the instance embeddings and the bag embeddings to be constrained by the fine-grained semantics, which is demonstrated by a theoretical analysis and significant improvement over the baseline.

Though self-supervised FGVC still exhibits an inferior performance to the state-of-the-art fully-supervised FGVC methods, its pre-training does not involve the fine-grained annotation. 
This property is particularly important for FGVC, as the fine-grained annotation demands expertise in fields such as biology, and is extremely difficult to collect. 
Therefore, self-supervised FGVC methods show great potential to alleviate the annotation labor, harness the huge amount of unlabeled data and enhance the robustness to unseen fine-grained data.
In the future, assume that millions of fine-grained images of the species can be collected, self-supervised pre-training is more effective to learn a discriminative fine-grained image representation. 
Thus, self-supervised FGVC methods hold importance in such scenarios and can be encouraged for more future exploration.

Concretely, our contribution can be summarized as follows:

\begin{itemize}
\item We propose a \textbf{C}ross-level \textbf{M}ulti-instance \textbf{D}istillation (CMD) framework for self-supervised fine-grained visual categorization. Multiple instance learning is introduced to model the relation between fine-grained semantics and image patches. 
\item We propose both intra-level and inter-level learning objectives for multi-instance distillation, which learn the relation between informative parts and fine-grained semantics among the image-level and region-level crops from both teacher and student net.
\item Experiments on CUB-200-2011, Stanford Cars and FGVC Aircraft show that the proposed CMD outperforms the contemporary method by up to 10.14\% and the state-of-the-art self-supervised representation learning methods by up to 19.78\% on top-1 accuracy.
\end{itemize}

The remainder of this paper is organized as follows. In Section~\ref{sec2}, the related work is summarized. 
In Section~\ref{sec3}, a detailed problem formulation of MIL and self-supervised FGVC is discussed. 
In Section~\ref{sec4}, the proposed CMD is presented in detail. 
In Section~\ref{sec5}, extensive experiments are implemented to validate the effectiveness of the proposed CMD. 
Finally, we draw the conclusion in Section~\ref{sec6}.

\section{Related work}
\label{sec2}
\subsection{Fine-grained Visual Categorization} 

Fine-grained visual categorization (FGVC) is a fundamental task for image understanding and pattern recognition.
It aims to discern a specific fine-grained category (\eg, golden retriever) from the subordinate coarse-grained category (\eg, dog).
The discriminative patterns among different fine-grained categories can be rather subtle, which are usually termed as parts.
They may only occupy several or tens of pixels in an entire image. 

Existing FGVC approaches are usually fully-supervised.
They follow the part-driven paradigm, which discerns the fine-grained semantics from the discriminative parts. 
The utilization of parts can be either explicit or implicit.
The explicit part selection pipelines are heavily inherited from the visual localization frameworks \cite{luo2019cross,ge2019weakly,han2022pcnn}. 
More recently, implicit part selection approaches have been extensively studies. 
Technically, these methods either build multi-scale representations \cite{zhang2019learning,touvron2019fixing,ji2020attention,du2020fine,zhao2021part} from CNN \cite{dubey2018pairwise,zhang2019learning,chang2023erudite} and ViT models \cite{zhao2021part,hu2021rams,he2021transfg}, or utilize attention modules \cite{ji2020attention,gao2020channel,zhuang2020learning,behera2021context}. 

On the other hand, there is a recent trend to combine FGVC with other tasks, such as object re-identification \cite{rao2021counterfactual,zhu2022dual}, image retrieval \cite{wei2021A2,shen2022semicon}, and multi-modal learning \cite{yang2022dynamic,wang2022knowledge}.
In general, enhancing the representation from subtle fine-grained patterns is able to also improve the discriminative ability of these tasks, yielding a better performance. 
However, how to distillate the key local patterns under multiple instance learning (MIL) formulation has not been explored by prior works, where the MIL formulation can pose a rigours constraint between the subtle patterns and fine-grained semantic categories.
Notice that, although several recent works leverage self-supervised techniques to refine the fine-grained patterns \cite{yang2022fine,Fang2024Contrast,ji2023siamese}, these methods still require the fine-grained annotations to supervise the model and thus are under the fully-supervised paradigm.

\subsection{Self-supervised Visual Representation Learning}

In the context of computer vision,
self-supervised learning (SSL) intends to learn a pre-text visual representation from unlabeled images for down-stream tasks \cite{rastegar2024learn}. 
Some representative works such as SimCLR \cite{chen2020simple}, MoCo \cite{chen2020improved}, BYOL \cite{grill2020bootstrap}, SimSiam \cite{chen2021exploring}, BarlowTwins \cite{zbontar2021barlow} and VICReg \cite{bardes2022vicreg}.
More recently, multiple advanced SSL methods have been proposed \cite{ben2023reverse,assran2023self,feng2023evolved}.
The overall idea of these methods is to learn an image representation from two or multiple augmented views.

On the other hand, self-distillation with no labels (DINO) shows its superiority \cite{caron2021emerging} against these image-level SSL methods.
Vision Transformer based DINO shows emerging properties to perceive more details in the image apart from the image-level semantics \cite{caron2021emerging}, which yield a stronger feature representation.
Later, its advanced version efficient self-supervised vision transformer (EsViT) is proposed to improve the computational efficiency and the view interaction \cite{li2021efficient}.
Compared with CNN based contrastive learning frameworks, ViT based DINO frameworks \cite{caron2021emerging,li2021efficient} are more capable to preserve the object structures in an image.
It motivates us to shift the focus from CNN backbone to ViT backbone for fine-grained self-supervised learning.
\textit{To the best of our knowledge}, learning fine-grained visual representation from self-supervised knowledge distillation remains unexplored.

\subsection{Self-supervised Fine-grained Visual Categorization}

Self-supervised FGVC is an emerging research topic, which intends to learn a pre-trained class-agnostic representation that can discriminate the fine-grained semantics. Pioneer works are summarized as follows.
Cole et al. empirically found some existing self-supervised learning pipelines suffered significant performance decline for FGVC task \cite{cole2022does}.
Shu et al. proposed a common rational learning strategy (LCR) for fine-grained visual categorization and retrieval \cite{shu2023learning}.
Kim et al. considered the fine-grained visual representation learning under a self-supervised open-set scenario \cite{kim2023coreset}.
Hu et al. proposed an asymmetric augmented self-supervised learning scheme, where the pre-trained representation can be used to query the fine-grained images \cite{hu2024asymmetric}.

On the other hand, as earlier FGVC pipelines adapt the localization pipeline for part proposals, in this paper, we consider the self-supervised localization approaches \cite{shu2022improving,zhao2021distilling,wu2021align,huang2022learning,peng2022crafting} as an alternative solution for self-supervised FGVC and make extensive comparison.
Compared with these methods, the proposed CMD does not reply on the part proposals as a prerequisite to learn fine-grained representation. Instead, it highlights the key instances implicitly in an one-stage self-supervised representation learning pipeline.

\subsection{Multiple Instance Learning}

Multiple instance learning (MIL) is a typical machine learning tool dealing with this so-called weak annotation scenario.
It models an image as a bag of instances with only a single bag annotation, which is able to find the key instances that trigger the bag annotation. 

In the deep learning era, MIL has been integrated into deep learning models in an end-to-end manner, which is termed as deep MIL.
Specifically, residual connection and deep supervision has turned out effective to improve the deep MIL features
\cite{wang2018revisiting,yan2018deep}.
Attention weights can also enhance the representation capability of deep MIL features \cite{ilse2018attention,bi2021local,yu2021mil}.

More recently, due to the rapid development of deep learning and computer vision, deep MIL has also demonstrated its representation learning ability on other tasks such as whole slide image classification \cite{wang2019bag,zhang2023multi}, aerial scene classification \cite{bi2020multiple,bi2022all}, weakly-supervised object detection \cite{tang2017deep,tang2017multiple} and weakly-supervised segmentation \cite{cheng2022sparse}.

\section{Preliminary}
\label{sec3}

\subsection{MIL Formulation} 

Given a bag $\mathbf{X}$ that consists of a set of instances $\mathbf{X}=\{\mathbf{x}^1, \cdots, \mathbf{x}^K\}$. 
The bag $\mathbf{X}$ corresponds to a binary label $Y \in \{0, 1\}$, while there is no label for each instance.

Multiple instance learning (MIL) assumes that each instance $\mathbf{x}^k$ ($t=1, \cdots, T$) exists a label $y^k \in \{0, 1\}$, but these instance-level labels remain unknown during the learning process.
It is formulated as
\begin{equation} 
\label{MILformulation}
Y=
\begin{cases}
0 & \text{if $\sum_{k=1}^K y^k$=0 }\\
1 & \text{else}
\end{cases}.
\end{equation} 

Avoiding the gradient vanishing problem is the key challenge to incorporate MIL into deep neural networks.
The bag label is assumed to be the Bernoulli distribution \cite{ilse2018attention,zaheer2017deep}, given by $Y\in[0,1]$.

\subsection{MIL Aggregation Function}

Assume that each instance $\mathbf{x}^k$ corresponds to an instance embedding $\mathbf{z}_{\mathbf{x}^k}$, and the bag $\mathbf{X}$ corresponds to a bag embedding $\mathbf{z}$.
The aggregation function $g(\cdot)$ bridges the gap between the instance embedding and bag embedding, by aggregating $\mathbf{z}_{\mathbf{x}^k}$ into $\mathbf{z}$. 
The aggregation function $g(\cdot)$ is assumed to be permutation invariant. 

\textbf{Lemma 1.} \textit{A scoring function for a set of instances $\mathbf{X}$, $S(X) \in \mathbb{R}$, is permutation invariant to the elements in $\mathbf{X}$, if and only if it can be decomposed in the following form:}
\begin{equation} 
\label{scorefunction}
S(X) = h (\sum_{\mathbf{x} \in \mathbf{X}} f(\mathbf{x})),
\end{equation}
\textit{where $f$ and $h$ are suitable transformations.} 

\textbf{Proof.} Please refer to \cite{zaheer2017deep} for the detailed proof. 

%Eq.~\ref{scorefunction}~indicates that, the transformation from instance to bag is a three-step procedure for MIL. 
%First step, a transformation $f$ converts the instances $\mathbf{x}^k$ to instance embedding $\mathbf{z}_{\mathbf{x}^k}$;
%Second step, an aggregation function $g(\cdot)$ converts the transformed instance embeddings in a permutation invariant manner;
%Third step, a transformation $h$ converts the aggregated representation into the bag score.

\iffalse
\subsection{Instance Paradigm v.s. Embedding Paradigm}

There are two domaint paradigms for existing deep multiple instance learning approaches, namely, instance-level paradigm and embedding-level paradigm, respectively.

\subsubsection{Instance-level Paradigm} 
The aggregation function $g(\cdot)$ directly transforms the instance embedding $\mathbf{z}_{\mathbf{x}^k}$ into the bag score. 
Consequently, the transformation $h(\cdot)$ degrades into an identity mapping. 

\subsubsection{Embedding-level Paradigm}
The function $f(\cdot)$ maps the
instances to a low-dimensional embedding $\mathbf{z}_{\mathbf{x}^k}$.
The aggregation function $g(\cdot)$ transforms $\mathbf{z}_{\mathbf{x}^k}$ into the bag embedding $\mathbf{z}$. Finally, the transformation $h(\cdot)$ transforms the bag embedding into the bag score.
\fi

\subsection{MIL Formulation for Self-supervised FGVC}

In self-supervised fine-grained visual representation learning task, there is neither explicit instance label $y^k$ nor bag label $Y$.
Following the MIL paradigm (Eq.~\ref{scorefunction}), the bag embedding is a pre-trained class-agnostic representation learned by a self-supervised framework.

Specifically, given two augmented views, following Eq.~\ref{scorefunction}, the bag embedding of each view, denoted as $\mathbf{z}_t$ and $\mathbf{z}_s$, can be generated.
Then, the MIL formulation under the self-supervised learning can be presented as 
\begin{equation} 
\label{selfbag}
\mathcal{L} (\mathbf{z}_t, \mathbf{z}_s),
\end{equation}
where $\mathcal{L}$ is a learning objective function to minimize the difference between $\mathbf{z}_t$ and $\mathbf{z}_s$.
For different self-supervised learning pipelines, the learning objective function $\mathcal{L}$ can be different. For example, for SimCLR \cite{chen2020simple}, it is the contrastive loss function. For Self-distillation with no labels (DINO) \cite{caron2021emerging}, it is a summed cross entropy loss function.

\section{Proposed Approach}
\label{sec4}

\begin{figure*}
    \centering
    \includegraphics[width=1.0\linewidth]{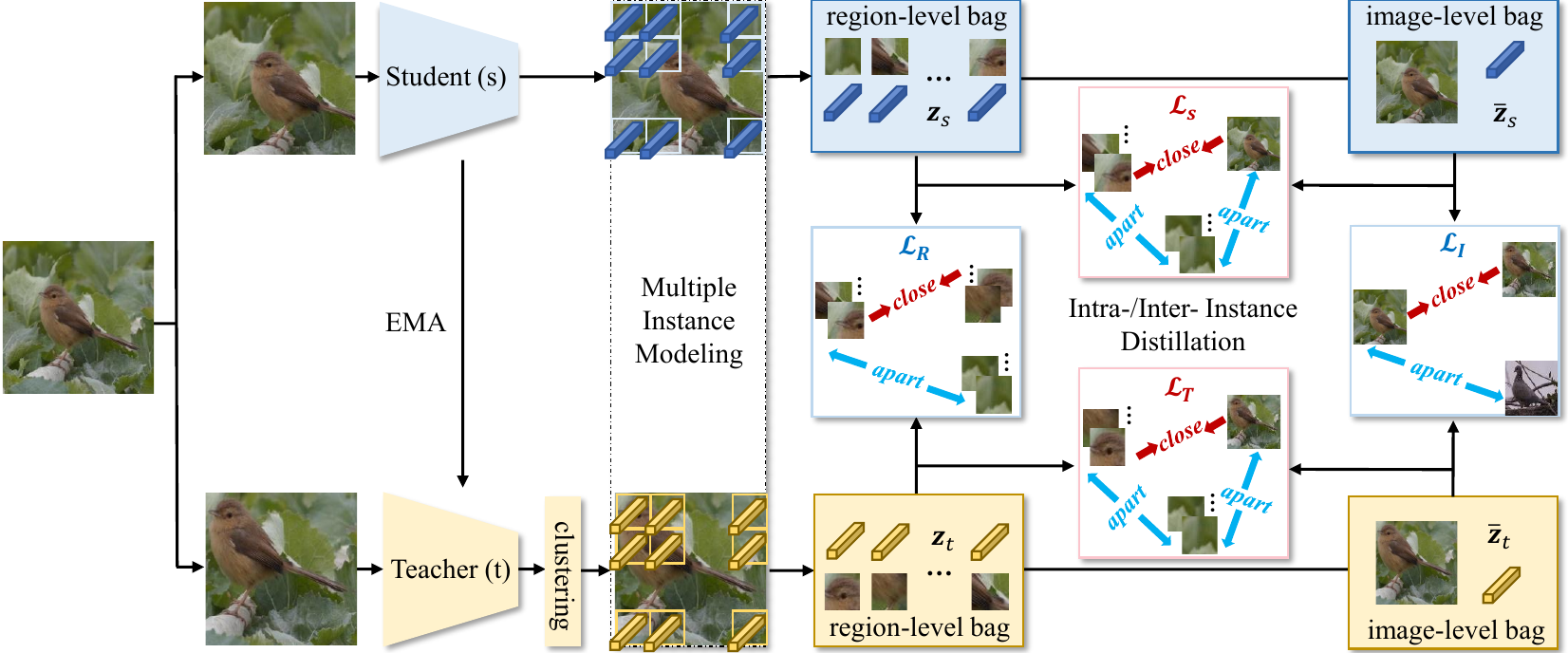}
    \vspace{-0.5cm}
    \caption{Framework overview of the proposed \textbf{C}ross-level \textbf{M}ulti-instance \textbf{D}istillation (CMD) for self-supervised fine-grained visual categorization. The proposed framework follows the self-distillation with no labels (DINO) \cite{caron2021emerging} paradigm. After feature extraction, three key steps are involved, namely, multi-instance modeling (Sec.~\ref{sec3.1}), intra-level multi-instance distillation (Sec.~\ref{sec3.2} with $\mathcal{L}_{I}$ and $\mathcal{L}_{R}$) and inter-level multi-instance distillation (Sec.~\ref{sec3.3} with $\mathcal{L}_{T}$ and $\mathcal{L}_{S}$). 
    }
    \label{framework}
\end{figure*}

Fig.~\ref{framework}~gives an overview of the proposed \textbf{C}ross-level \textbf{M}ulti-instance \textbf{D}istillation (CMD) for self-supervised FGVC.
On top of the Self-distillation with no labels (DINO) \cite{caron2021emerging} paradigm,
three key steps are involved in CMD, namely, multi-instance modeling (in Sec.~\ref{sec3.1}), intra-level multi-instance distillation (in Sec.~\ref{sec3.2}) and inter-level multi-instance distillation (in Sec.~\ref{sec3.3}). 
Finally, more implementation details and loss functions are provided in Sec.~\ref{sec3.4}.

The proposed CMD is distinct from the common patch- and image- level losses in multiple aspects. Firstly, our integration of patch and image-level losses is under the rigorous multiple instance learning (MIL) formulation, which provides an implicit constraint of the fine-grained semantics on both region- and image- level crops. Secondly, directly applying patch- and image- level losses could not resolve the learning objective inconsistency problem in self-supervised FGVC.
CMD innovatively resolves it by intra- and inter- level distillation over both patch- and image- level losses.
It allows self-supervised learning to be optimized by the patch-level objective while at the same time be constrained by the image-level context. Thirdly, CMD makes a new exploration to leverage both patch- and image- level information for the DINO paradigm, and to advance MIL for both patch- and image- level information under self-supervised learning.

\subsection{Multi-instance Modeling}
\label{sec3.1}
\subsubsection{Motivation} Self-supervised FGVC needs to bridge the relation between several highlighted key patches of the image and the fine-grained semantics.
To this end, MIL is introduced to model each patch of the image as an instance and the overall image as a bag. 
Although MIL has demonstrated great success in weakly-supervised learning scenarios, some unique challenges remain in our task.

\subsubsection{Challenges} Usually, there is bag-level annotation under MIL formulation \cite{wang2013max,wang2015relaxed,tang2017multiple,tang2016learning}. 
However, for self-supervised FGVC, there is neither bag-level nor instance-level annotation. 
Thus, the bag representation has to be learned in a self-supervised manner.
In addition, to comprehensively exploit the relation between key patches and fine-grained semantics, it is necessary to inherit both region-level and image-level crops in existing SSL pipelines \cite{li2021efficient}. 
For simplicity, here we model both region-level and image-level crops as \textit{bag}.

\subsubsection{Notations \& Definitions} Given an image $\mathbf{X}$ as input, and two augmented views $\mathbf{X}_t$ and $\mathbf{X}_s$ for the teacher net $t$ and student net $s$, respectively. 
For both nets, some crops are image-level (denoted as $\mathcal{I}$) and some crops are region-level (denoted as $\mathcal{R}$).
For convenience, we formulate an instance $\mathbf{x}_s^{i}$ and $\mathbf{x}_t^{i}$ from an augmented view $\mathbf{X}_s$ and $\mathbf{X}_t$ is an image patch of the same size as the patch size in a Vision Transformer backbone, denoted as $\mathbf{X}_s=\{\mathbf{x}_s^{1}, \cdots, \mathbf{x}_s^{i}, \cdots\}$ and $\mathbf{X}_t=\{\mathbf{x}_t^{1}, \cdots, \mathbf{x}_t^{i}, \cdots\}$.
For both teacher and student net, the Swin Transformer \cite{liu2021swin} is used as the default backbone for feature extraction.
A linear layer $h(\cdot)$ is used to transform the patch-wise embedding $\mathbf{R}_{t}$ and $\mathbf{R}_{s}$ into a set of instance embedding $\mathbf{z}_{\mathbf{x}_t^i} \in \mathbb{R}^{(w \cdot h) \times 1}$ or $\mathbf{z}_{\mathbf{x}_s^i} \in \mathbb{R}^{(w \cdot h) \times 1}$.

\subsubsection{Multiple Instance Aggregation Function} The key to model the relation between instance and bag is the aggregation function, which has to be permutation invariant to the order change of instances \cite{ilse2018attention,wang2018revisiting}. 
Take the teacher net as an example, the mean pooling based aggregation function is used to learn both the region-level bag embedding $\mathbf{z}_{t} \in \mathbb{R}^{(w \cdot h) \times 1}$ and image-level bag embedding $\mathbf{\overline{z}}_{t} \in \mathbb{R}^{(w \cdot h) \times 1}$, given by
\begin{equation} 
\label{regionbagemb}
\mathbf{z}_{t}=\frac{1}{\mathcal{R}} \sum_{\mathbf{x}_{t}^{i} \in \mathcal{R}} \mathbf{z}_{\mathbf{x}_{t}^{i}}, \quad
\mathbf{\overline{z}}_{t}=\frac{1}{\mathcal{I}} \sum_{\mathbf{x}_{t}^{i} \in \mathcal{I}} \mathbf{z}_{\mathbf{x}_{t}^{i}}.
\end{equation}

For student network, the aggregation from the instance representation to the bag representation is the same as the above process, and we get both crop-level bag representation $\mathbf{z}_{s} \in \mathbb{R}^{(w \cdot h) \times 1}$ and image-level bag representation $\mathbf{\overline{z}}_{s} \in \mathbb{R}^{(w \cdot h) \times 1}$.

\subsubsection{Permutation Invariance Property}

Here we provide the proof that, in the proposed framework, the aggregation function from both region-level and image-level crops to bag embedding meets the permutation invariance property, which is both sufficient and necessary to hold the multiple instance formation.

\textbf{Theorem 1.} 
\textit{The transformation from instance embedding to bag embedding of the region-crop (Eq.~\ref{regionbagemb}) is permutation-invariant.}

\textbf{Proof.} Take the teacher net $t$ as an example. In the proposed framework, the transformation between instance embedding $\mathbf{z}_{\mathbf{x}_t}$ and bag embedding $\mathbf{z}_{t}$ of a region crop $\mathcal{R}$ is calculated as
\begin{equation} 
\label{regionbagemb}
\mathbf{z}_{t}=\frac{1}{\mathcal{R}} \sum_{\mathbf{x}_{t}^{k} \in \mathcal{R}} \mathbf{z}_{\mathbf{x}_{t}^{k}},
\end{equation}
where we have $\mathcal{R}=\{\mathbf{x}_{t}^{1}, \cdots, \mathbf{x}_{t}^{k}, \cdots, \mathbf{x}_{t}^{R}\}$, and the instances $\{\mathbf{x}_{t}^{1}, \cdots, \mathbf{x}_{t}^{k}, \cdots, \mathbf{x}_{t}^{R}\}$ correspond to the instance embeddings $\{ \mathbf{z}_{\mathbf{x}_{t}^{1}}, \cdots, \mathbf{z}_{\mathbf{x}_{t}^{k}}, \cdots, \mathbf{z}_{\mathbf{x}_{t}^{R}} \}$.

Assume $\{\mathbf{x}_{t}^{\sigma{(1)}}, \cdots, \mathbf{x}_{t}^{\sigma{(k)}}, \cdots, \mathbf{x}_{t}^{\sigma{(R)}}\}$ is an arrangement of $\{\mathbf{x}_{t}^{1}, \cdots, \mathbf{x}_{t}^{k}, \cdots, \mathbf{x}_{t}^{R}\}$, which corresponds to the instance embeddings $\{ \mathbf{z}_{\mathbf{x}_{t}^{\sigma{(1)}}}, \cdots, \mathbf{z}_{\mathbf{x}_{t}^{\sigma{(k)}}}, \cdots, \mathbf{z}_{\mathbf{x}_{t}^{\sigma{(R)}}} \}$.

A permutation transformation $t$ of the instances is a process presented as
\begin{equation} 
\label{regiontrans}
t: \{ \mathbf{x}_{t}^{1}, \cdots, \mathbf{x}_{t}^{k}, \cdots, \mathbf{x}_{t}^{R} \} \rightarrow \{ \mathbf{x}_{t}^{\sigma{(1)}}, \cdots, \mathbf{x}_{t}^{\sigma{(k)}}, \cdots, \mathbf{x}_{t}^{\sigma{(R)}} \}. 
\end{equation}

Obviously, we have
\begin{equation} 
\label{regionproof}
\mathbf{z}_{t}=\frac{1}{\mathcal{R}} \sum_{\mathbf{x}_{t}^{k} \in \mathcal{R}} \mathbf{z}_{\mathbf{x}_{t}^{k}}=\frac{1}{\mathcal{R}} \sum_{\mathbf{x}_{t}^{\sigma{(k)}} \in \mathcal{R}} \mathbf{z}_{\mathbf{x}_{t}^{\sigma{(k)}}}.
\end{equation}

\textbf{Theorem 2.} 
\textit{The transformation from instance embedding to bag embedding of the image-crop (Eq.~\ref{regionbagemb}) is permutation-invariant.}

\textbf{Proof.} The proof is similar to the proof of Theorem 1. 
Still take the teacher net $t$ as an example. In the proposed framework, the transformation between instance embedding $\mathbf{z}_{\mathbf{x}_t}$ and bag embedding $\mathbf{\overline{z}}_{t}$ of a region crop $\mathcal{I}$ is calculated as
\begin{equation} 
\label{imagebagemb}
\mathbf{\overline{z}}_{t}=\frac{1}{\mathcal{I}} \sum_{\mathbf{x}_{t}^{k} \in \mathcal{I}} \mathbf{z}_{\mathbf{x}_{t}^{k}},
\end{equation}
where we have $\mathcal{R}=\{\mathbf{x}_{t}^{1}, \cdots, \mathbf{x}_{t}^{k}, \cdots, \mathbf{x}_{t}^{I}\}$, and the instances $\{\mathbf{x}_{t}^{1}, \cdots, \mathbf{x}_{t}^{k}, \cdots, \mathbf{x}_{t}^{I}\}$ correspond to the instance embeddings $\{ \mathbf{z}_{\mathbf{x}_{t}^{1}}, \cdots, \mathbf{z}_{\mathbf{x}_{t}^{k}}, \cdots, \mathbf{z}_{\mathbf{x}_{t}^{I}} \}$.

Assume $\{\mathbf{x}_{t}^{\sigma{(1)}}, \cdots, \mathbf{x}_{t}^{\sigma{(k)}}, \cdots, \mathbf{x}_{t}^{\sigma{(I)}}\}$ is an arrangement of $\{\mathbf{x}_{t}^{1}, \cdots, \mathbf{x}_{t}^{k}, \cdots, \mathbf{x}_{t}^{I}\}$, which corresponds to the instance embeddings $\{ \mathbf{z}_{\mathbf{x}_{t}^{\sigma{(1)}}}, \cdots, \mathbf{z}_{\mathbf{x}_{t}^{\sigma{(k)}}}, \cdots, \mathbf{z}_{\mathbf{x}_{t}^{\sigma{(I)}}} \}$.

A permutation transformation $t$ of the instances is a process presented as
\begin{equation} 
\label{imagetrans}
t: \{ \mathbf{x}_{t}^{1}, \cdots, \mathbf{x}_{t}^{k}, \cdots, \mathbf{x}_{t}^{I} \} \rightarrow \{ \mathbf{x}_{t}^{\sigma{(1)}}, \cdots, \mathbf{x}_{t}^{\sigma{(k)}}, \cdots, \mathbf{x}_{t}^{\sigma{(I)}} \}. 
\end{equation}

Obviously, we have
\begin{equation} 
\label{imageproof}
\mathbf{\overline{z}}_{t}=\frac{1}{\mathcal{I}} \sum_{\mathbf{x}_{t}^{k} \in \mathcal{I}} \mathbf{z}_{\mathbf{x}_{t}^{k}}=\frac{1}{\mathcal{I}} \sum_{\mathbf{x}_{t}^{\sigma{(k)}} \in \mathcal{I}} \mathbf{z}_{\mathbf{x}_{t}^{\sigma{(k)}}}.
\end{equation}

\subsection{Intra-level Multi-instance Distillation}
\label{sec3.2}

\subsubsection{Motivation} In the feature space, the distance between the region-level bag representation from teacher net $\mathbf{z}_{t}$ and from student net $\mathbf{z}_{s}$, and the distance between the image-level bag representation from teacher net $\mathbf{\overline{z}}_{t}$ and from student net $\mathbf{\overline{z}}_{s}$,
both needs to be minimized.
In this way, the cropped image
pairs / cropped region pairs that represent similar fine-grained categories are close together, while those represent dissimilar fine-grained categories are far apart.

The intra-level multi-instance distillation is proposed to realize the above two objectives, which distillates image-level crops (Eq.~\ref{imagelevelloss}) and region-level crops (Eq.~\ref{regionlevelloss}), respectively.

\subsubsection{Notations \& Definitions} For both teacher net $t$ and student net $s$, after view augmentation, there are multiple view crops (both region crops and image crops) from an image, denoted as $\mathbf{V}=\{\mathbf{X}_t\}$ and $\mathbf{V}^{*}=\{\mathbf{X}_s\}$, respectively.
Assume that the crop pairs from the teacher and student net are organized as $\mathcal{P}=\{(s,t)| \mathbf{X}_s \in \mathbf{V}, \mathbf{X}_t \in \mathbf{V}^{*}, s \neq t \}$.

\subsubsection{Image-level Multi-instance Distillation} 
After modeled by MIL, the image-level bag representation from teacher net $\mathbf{\overline{z}}_{t}$ and student net $\mathbf{\overline{z}}_{s}$ is minimized pair-by-pair, given by
\begin{equation} 
\label{imagelevelloss}
    \mathcal{L}_{\mathcal{I}}=-\frac{1}{\mathcal{P}} \sum_{(s,t) \in \mathcal{P}} \mathbf{\overline{z}}_s log \mathbf{\overline{z}}_t.
\end{equation}

\subsubsection{Region-level Multi-instance Distillation} 

Similarly, after modeled by MIL, the region-level bag representation from teacher net $\mathbf{z}_{t}$ and from student net $\mathbf{z}_{s}$ is minimized firstly region-by-region for each pair $(t,s)$ and then averaged pair-by-pair, given by
\begin{equation} 
\label{regionlevelloss}
    \mathcal{L}_{\mathcal{R}}=- \frac{1}{\mathcal{P}} \sum_{(s,t) \in \mathcal{P}} \frac{1}{T} \sum_{i=1}^{T} \mathbf{z}_{j^{*}} log \mathbf{z}_i,
\end{equation}
where
$j^{*}$ is the index of the feature in $\mathbf{R}_{t}$ that has the highest similarity with the $i^{th}$ feature in $\mathbf{R}_{s}$ (defined in Sec.~\ref{sec3.1}), and $T$ is the length of the patch-wise embedding
\begin{equation} 
\label{ijcos}
j^{*} = arg \mathop{max}\limits_{j} \frac{R_{i}^{T} R_{j}}{|R_i| |R_j|},
\end{equation}
where $z_i=h(R_i)$ and $z_j=h(R_j)$, and $R_{i} \in \mathbf{R}_{s}$, $R_{j} \in \mathbf{R}_{t}$.

\subsection{Inter-level Multi-instance Distillation}
\label{sec3.3}

\subsubsection{Motivation} 
The intra-level multi-instance distillation (in Sec.~\ref{sec3.2}) does not take the semantic relation between the region-level crops and image-level crops from the same augmented view into account. 
More abundant information to describe the key regions can rest in the relation between region-level and image-level crops from the same augmented view.
Ideally, in our framework, the fine-grained semantics should be consistent between both image-level and region-level crops.

To this end, we further propose the inter-level multi-instance distillation to exploit this aspect. 
It helps the cropped image-region pairs that represent similar fine-grained categories are close together, while those represent dissimilar fine-grained categories are far apart.

\subsubsection{Notations \& Definitions} 
Still consider the crops $\mathbf{X}_s \in \mathbf{V}$ inside the student net and the crops $\mathbf{X}_t \in \mathbf{V}^{*}$ inside the teacher net. 
These crops $\mathbf{X}_s \in \mathbf{V}$ from the student net organize a crop pair as $\mathcal{V}=\{(i,j)| \mathbf{X}_s^{i} \in \mathbf{V}, \mathbf{X}_s^{j} \in \mathbf{V}, i \neq j \}$, where $\mathbf{X}_s^{i}$, $\mathbf{X}_s^{j}$ correspond to either a region-level bag embedding $\mathbf{z}_s^{i}$, $\mathbf{z}_s^{j}$ or an image-level bag embedding $\mathbf{\overline{z}}_{s}$, $\mathbf{\overline{z}}_{t}$.

\subsubsection{Inter-level Multi-instance Distillation of Student Net} 
For student net $s$, the embedding between each crop pair, either image-level or region-level, is minimized by the
Kullback–Leibler divergence \cite{hall1987kullback}, given by
\begin{equation} 
\footnotesize
\label{stuinterloss}
    \mathcal{L}_{\mathcal{S}}= \frac{1}{\mathcal{V}} \sum_{(i,j) \in \mathcal{V}} (\mathbf{z}_{s}^{i} log \frac{\mathbf{z}_{s}^{i}}{\mathbf{z}_{s}^{j}}+ 
    \mathbf{z}_{s}^{i} log \frac{\mathbf{z}_{s}^{i}}{\mathbf{\overline{z}}_{s}^{j}}+
    \mathbf{\overline{z}}_{s}^{i} log \frac{\mathbf{\overline{z}}_{s}^{i}}{\mathbf{z}_{s}^{j}}+
    \mathbf{\overline{z}}_{s}^{i} log \frac{\mathbf{\overline{z}}_{s}^{i}}{\mathbf{\overline{z}}_{s}^{j}}
    ).
\end{equation}

\subsubsection{Inter-level Multi-instance Distillation of Teacher Net}
Similarly, for the teacher net $t$, we have the crop pair $\mathcal{V}^{*}=\{(i,j)| \mathbf{X}_t^{i} \in \mathbf{V}^{*}, \mathbf{X}_t^{j} \in \mathbf{V}^{*}, i \neq j \}$, where $\mathbf{X}_t^{i}$, $\mathbf{X}_t^{j}$ correspond to either a region-level bag embedding $\mathbf{z}_s^{i}$, $\mathbf{z}_s^{j}$ or an image-level bag embedding $\mathbf{\overline{z}}_{s}$, $\mathbf{\overline{z}}_{t}$.
%Also for simplicity, here $\mathbf{z}_t$ refers to the bag representation regardless of region-level or image-level.
The learning objective of the inter-level multi-instance distillation for the teacher net is also measured by Kullback–Leibler divergence \cite{hall1987kullback}, and 
can be computed as
\begin{equation}
\footnotesize
\label{teainterloss}
    \mathcal{L}_{\mathcal{T}}= \frac{1}{\mathcal{V}^{*}} \sum_{(i,j) \in \mathcal{V}^{*}} (\mathbf{z}_{t}^{i} log \frac{\mathbf{z}_{t}^{i}}{\mathbf{z}_{t}^{j}}+ 
    \mathbf{z}_{t}^{i} log \frac{\mathbf{z}_{t}^{i}}{\mathbf{\overline{z}}_{t}^{j}}+
    \mathbf{\overline{z}}_{t}^{i} log \frac{\mathbf{\overline{z}}_{t}^{i}}{\mathbf{z}_{t}^{j}}+
    \mathbf{\overline{z}}_{t}^{i} log \frac{\mathbf{\overline{z}}_{t}^{i}}{\mathbf{\overline{z}}_{t}^{j}}).
\end{equation}

\subsection{Loss and Implementation Details}
\label{sec3.4}

\begin{algorithm}[!t]
\caption{Pseudo code of the proposed Cross-level Multi-instance Distillation, PyTorch Style }
\label{proposedalgo}
\begin{algorithmic}[1]
%\begin{lstlisting}[style=Python3]
\State $t$, $s$: teacher and student net; $h(\cdot)$: linear layer
\State $\mathbf{X}$, $\mathbf{X}_v$, $\mathbf{X}_r$: origin input and two augmented views
\State $Ci$, $Cr$: view and region center
\State
\State t.params = s.params
\State \textcolor{blue}{\textbf{for}} $\mathbf{X}$ \textbf{in loader}:
\State \quad $\mathbf{X}_{1}$, $\mathbf{X}_{2}$ = augment($\mathbf{X}$), augment($\mathbf{X}$) 
%\quad  \textcolor{YellowGreen}{\#two views from an image}
\State \quad $\mathbf{R}_{t}$, $\mathbf{R}_{s}$ = t($\mathbf{X}_{1}$), s($\mathbf{X}_{2}$) 
\State \quad $\mathbf{z}_{x_{t}}$, $\mathbf{z}_{x_{s}}$ = h($\mathbf{R}_{t}$), h($\mathbf{R}_{s}$) 
%\State
\State \quad for $\mathbf{z}_{x_{s}}, \mathbf{z}_{x_{t}}$ in $\mathcal{R}$: \quad \textcolor{YellowGreen}{\# bag embedding for region crops}
\State \quad \quad \quad $\mathbf{z}_{s}$ = mean(\{$\mathbf{z}_{x_{s}}$\}), $\mathbf{z}_{t}$ = mean(\{$\mathbf{z}_{x_{t}}$\})
%\State
\State \quad for $\mathbf{z}_{x_{s}}$, $\mathbf{z}_{x_{t}}$ in $\mathcal{I}$: \quad \textcolor{YellowGreen}{\# bag embedding for image crops}
\State \quad \quad \quad $\overline{\mathbf{z}_{s}}$ = mean(\{$\mathbf{z}_{x_{s}}$\}), $\overline{\mathbf{z}_{t}}$ = mean(\{$\mathbf{z}_{x_{t}}$\})
%\State
\State \quad \textcolor{YellowGreen}{\# Intra-level Multi-instance Distillation}
\State \quad $\mathcal{L}_{I}$ = $H_{I}$($\overline{\mathbf{z}_{t}}$, $\overline{\mathbf{z}_{s}}$) \quad \textcolor{YellowGreen}{\# for image-level crops}
\State \quad $\mathcal{L}_{R}$ = $H_{R}$($\mathbf{z}_{t}$, $\mathbf{z}_{s}$, $\mathbf{R}_{t}$, $\mathbf{R}_{s}$) \quad \textcolor{YellowGreen}{\# for region-level crops}
%\State
\State \quad \textcolor{YellowGreen}{\# Inter-level Multi-instance Distillation}
\State \quad $\mathcal{L}_{T}$=$H_{T}$($\mathbf{z}_{t}$, $\overline{\mathbf{z}_{t}}$) \quad \textcolor{YellowGreen}{\# for region-image crops in $t$}
\State \quad $\mathcal{L}_{S}$=$H_{S}$($\mathbf{z}_{s}$, $\overline{\mathbf{z}_{s}}$) \quad \textcolor{YellowGreen}{\# for region-image crops in $s$}
\State
\State \quad $\mathcal{L}$ = ($\mathcal{L}_{I}$ + $\mathcal{L}_{R}$)/2 + $\lambda$ ($\mathcal{L}_{T}$ + $\mathcal{L}_{S}$) \quad \textcolor{YellowGreen}{\# total loss}
\State \quad loss.backward()
%\State
\State \quad update(s) \quad  \textcolor{YellowGreen}{\# update student, teacher and centers}
\State \quad t.params = a * t.params + (1-a) * s.params  \quad \textcolor{YellowGreen}{\# EMA}
\State \quad  Ci = b * Ci + (1-b) * cat([$\mathbf{R}_t$, $\mathbf{R}_s$].mean(0)) 
\State \quad \textcolor{YellowGreen}{\# EMA for image center}
\State \quad  Cr = b * Cr + (1-b) * cat([$\mathbf{z}_t$, $\mathbf{z}_s$].mean(0)) 
\State \quad \textcolor{YellowGreen}{\# EMA for region center}
\State
\State \textcolor{blue}{\textbf{def}}  \textbf{$H_I$(s, t)}:
\State  \quad   t = t.detach() \quad  \textcolor{YellowGreen}{\# stop gradient}
\State  \quad   s = softmax(s / tmp\_s, dim=-1)
\State  \quad   t = softmax((t - Ci) / tmp\_t, dim=-1)
\State  \quad   \textcolor{blue}{\textbf{return}} - (t * log(s)).sum(dim=-1).mean()
\State
\State  \textcolor{blue}{\textbf{def}} $H_R$($\mathbf{R}_s$, $\mathbf{R}_t$, $\mathbf{z}_s$, $\mathbf{z}_t$):
\State  \quad     $\mathbf{R}_t$ = $\mathbf{R}_t$.detach() \quad \quad \quad \textcolor{YellowGreen}{\# stop gradient}
\State  \quad     $\mathbf{R}_s$ = softmax($\mathbf{R}_s$ / tmp\_s, dim=-1) \quad \quad \quad %\textcolor{YellowGreen}{\# n×T×K}
\State  \quad     $\mathbf{R}_t$ = softmax($\mathbf{R}_t$ - Cr) / tmp\_t, dim=-1) \quad \quad %\quad \textcolor{YellowGreen}{\# n×T×K}
\State  \quad     sim\_matrix = torch.matmul($\mathbf{z}_s$, $\mathbf{z}_t$.permute(0, 2, 1)) %\quad \quad \quad \textcolor{YellowGreen}{\# n×T×T}
\State \quad     sim\_idx = sim\_matrix.max(dim=-1)[1].unsqueeze(2) 
%\quad \quad \quad \textcolor{YellowGreen}{\# n×T×1}
\State  \quad     $\mathbf{R}_t$\_idxed = torch.gather($\mathbf{R}_t$, 1, sim\_idx.expand(-1, -1, pt.size(2)))
\State  \quad    \textcolor{blue}{\textbf{return}} - ($\mathbf{R}_t$\_idxed * log($\mathbf{R}_s$)).sum(dim=-1).mean()
\State
\State  \textcolor{blue}{\textbf{def}} $\mathbf{H}_T$($\mathbf{z}$, $\overline{\mathbf{z}}$):
\State  \quad \quad     \textcolor{blue}{\textbf{return}} F.kl\_div($\mathbf{z}$, $\overline{\mathbf{z}}$)
\State
\State  \textcolor{blue}{\textbf{def}} $\mathbf{H}_S$($\mathbf{z}$, $\overline{\mathbf{z}}$):
\State  \quad \quad    \textcolor{blue}{\textbf{return}} F.kl\_div($\mathbf{z}$, $\overline{\mathbf{z}}$)
\end{algorithmic}
\end{algorithm}

The overall learning objective of the proposed CMD framework is a combination of the intra-level and inter-level multi-instance distillation, computed as
\begin{equation} 
\label{totalloss}
    \mathcal{L} = (\mathcal{L}_{\mathcal{I}} + \mathcal{L}_{\mathcal{R}})/2 + \lambda_1 (\mathcal{L}_{\mathcal{S}}+\mathcal{L}_{\mathcal{T}}).
\end{equation}
where $\lambda$ is a loss weight to balance the impact between the intra- and inter- level multi-instance distillation. Empirically, we set $\lambda_1$=0.1.
To intuitively demonstrate the specific steps and the implementation of the proposed CMD, a PyTorch-style algorithm is attached in Algorithm~\ref{proposedalgo}.

Following the convention of recent self-distillation with no labels paradigm (\eg DINO \cite{caron2021emerging}, EsViT \cite{li2021efficient}), the parameters of teacher net and the student net are updated in an alternative manner. 
Given a frozen teacher net, the student net is updated by minimizing its full learning objective as $\mathbf{\theta}_s \leftarrow {\rm arg min}_{\mathbf{\theta}_s} \mathcal{L}(s, t; \mathbf{\theta}_s)$. 
Instead, the teacher net is updated by using an exponential moving average (EMA) from the weights of the student net, presented as
$\mathbf{\theta}_t \leftarrow \lambda \mathbf{\theta}_t + (1-\lambda) \mathbf{\theta}_s$.
Here $\lambda$ is a weight parameter that is parameterized by a cosine schedule, and varies from 0.996 to 1. 

By default, both the teacher net and the student net have 2 image-level crops and 8 region-level crops. The image-level crops have a size of $224\times224\times3$, the region-level crops have a size of $96\times96\times3$, and each instance for multiple instance modeling has a size of $32\times32\times3$.
Consequently, each image-level crop has 49 instances, and each region-level crop has 9 instances.

Both the teacher net and the student net use the Swin-Tiny backbone \cite{liu2021swin} as the feature extractor, and the pre-trained weights from ImageNet are used as the initial weights. 
To train the entire framework, the initial learning rate is $5\times10^{-4}$, and the weight decay is set 0.05.
The AdamW optimizer is utilized and the model training terminates after 300 epochs.

\subsection{Theoretical Analysis}
\label{sec3.5}

Overall, SSL uses image-level learning objective, but FGVC requires region-level learning objective to discern tiny discriminative patterns, which easily leads to the \textit{ambiguous gradient} problem.
This subsection provides a conceptual and high-level discussion of this problem. 

Given a dataset of $N$ unlabeled samples. After feature extraction, assume each sample is represented by $\mathbf{x}_1$,  $\mathbf{x}_2$, $\ldots$, $\mathbf{x}_N$, where $i$ is the $i$-th samples, and $\mathbf{x}_i \in \mathbb{R}^{1 \times (w\cdot h \cdot c)}$. Here $(w\cdot h \cdot c)$ proportional to the width, height and channel of deep features, and are high dimensional.

For conventional SSL (e.g., SimCLR, MoCo), the learning objective $\mathcal{L}_{SSL}$ is to minimize the distance between $\mathbf{x}_i$ and a sample $\mathbf{x}^+$ of the same category, while at the same time maximize the distance between $\mathbf{x}_i$ and a sample $\mathbf{x}^-$ of a different category, given by
\begin{equation} 
\label{sslexist}
%\footnotesize
\mathcal{L}_{SSL} = \sum_i^N  (\rm{min} \quad Dist(\mathbf{x}_i, \mathbf{x}^+) + \rm{max} \quad Dist(\mathbf{x}_i, \mathbf{x}^-) ),
\end{equation} 
where ${\rm Dist}$ is the distance metric, e.g., $l$-2 loss, KL-loss. 

For FGVC, still consider the above dataset.
The difference is, these $N$ samples are from the same coarse-category (e.g., dog), and they have much closer distance in the feature space.

The discern between fine-grained categories is to use tiny and subtle feature $\mathbf{x}_i^{1} \in \mathbb{R}^{1 \times (w_1 \cdot h_1 \cdot c)}$ from $\mathbf{x}_1 \in \mathbb{R}^{1 \times (w \cdot h \cdot c)}$ for categorization, where only a minor $w_1$ and $h_1$ from the width and height dimensions are useful, and $w_1 \cdot h_1 \ll w \cdot h$.
Let we denote the rest feature as $\mathbf{x}_i^{2} \in \mathbb{R}^{1 \times ((w-w_1) \cdot (w-h_1) \cdot c)}$.

In \textit{image space}, given an $256\times256$ image, the fine-grained patterns may only occupies tens of pixels (e.g., 50).
In \textit{feature space}, when both width and height are down-sampled (e.g., into a quarter), the fine-grained patterns $\mathbf{x}_i^{1}$ may only occupies a handful of pixels (e.g., 3 pixels), while the overall feature map $\mathbf{x}_i$ has a size of $64\times64$. The estimation in \textit{feature space} is $(w_1 \cdot h_1) / (w \cdot h) \approx 3/(64\times64) \approx 7.3 \times 10^{-4}$.

When using conventional SSL for FGVC, it implements Eq.1 where $\mathbf{x}_i^{1}$ has dominant impact.
The gradient is:
\begin{equation} 
\begin{aligned}
\label{sslgra}
\frac{\partial \mathcal{L}_{SSL}}{\partial \mathbf{x}_i} & = \frac{\partial \mathcal{L}_{SSL}}{\partial \mathbf{x}_i^1} \cdot \frac{\partial  \mathbf{x}_i^1}{\partial \mathbf{x}_i} \\ & = \frac{\partial (\rm{min} \quad Dist(\mathbf{x}_i, \mathbf{x}^+))}{\partial \mathbf{x}_i^1} \cdot \frac{\partial  \mathbf{x}_i^1}{\partial \mathbf{x}_i} \\ & + \frac{\partial (\rm{max} \quad Dist(\mathbf{x}_i, \mathbf{x}^-))}{\partial \mathbf{x}_i^1} \cdot \frac{\partial  \mathbf{x}_i^1}{\partial \mathbf{x}_i},
\end{aligned}
\end{equation} 

The dramatic ration $(w_1 \cdot h_1) / (w \cdot h)$ leads to a constant value in most of the channels when calculating the gradient $\frac{\partial \mathbf{x}_i^{1}}{\partial \mathbf{x}_i}$.
In fact, according to the above estimation, only 0.07\% channels are informative for fine-grained patterns.
It is more likely to cause ambiguous gradient, yielding a less discriminative feature space for FGVC.

In contrast, we formulate each image patch as an instance to better discern fine-grained pattern $\mathbf{x}_i^1$. In our proposed CMD, it is not based on the sample-wise learning objective, but instance-wise learning objective, given by:

\begin{equation} 
\mathcal{L}_{CMD}= \sum_i^N (\rm{min} \quad Dist(\mathbf{x}_i^1, \mathbf{x}^{1+}) + \rm{max} \quad Dist(\mathbf{x}_i^2, \mathbf{x}^{2-}) ),
\end{equation}

Consequently, the impact of fine-grained patterns $\mathbf{x}_i^1$ is:

\begin{equation} 
\frac{\partial \mathcal{L}_{CMD}}{\partial \mathbf{x}_i}= \frac{\partial \rm{min} \quad Dist(\mathbf{x}_i^1, \mathbf{x}^{1+})}{\partial \mathbf{x}_i^1}.
\end{equation}

The negligible term $\frac{\mathbf{x}_i^{1}}{\partial \mathbf{x}_i}$ that leads to \textit{ambiguous gradients} is eliminated in the proposed CMD, which helps learn a more discriminative FGVC feature space.
In addition, the calculation of $\frac{\partial \mathbf{x}_i^{1}}{\partial \mathbf{x}_i}$ can be specified by the distance function $\rm{Dist}$ of a specified SSL pipeline.

\section{Experimental Analysis}
\label{sec5}

\subsection{Dataset \& Evaluation Protocols}

\textbf{CUB-200-2011}\footnote{\url{https://www.vision.caltech.edu/datasets/cub_200_2011/}} is a FGVC dataset that contains 200 bird species. It has 5,994 training samples and 5,794 test samples. 

\textbf{Stanford Cars} \footnote{\url{https://ai.stanford.edu/~jkrause/cars/car_dataset.html}} contains 196 fine-grained categories, with 8,144 training samples and 8,041 test samples.

\textbf{FGVC Aircraft} \footnote{\url{https://www.robots.ox.ac.uk/~vgg/data/fgvc-aircraft/}} contains 102 fine-grained aircraft categories. It has a total number of 6,667 training samples and 3,333 test samples.

\textbf{NA-BIRDS} \footnote{\url{https://dl.allaboutbirds.org/nabirds}}
contains 555 fine-grained bird species. It has a total number of 23,929 training samples and 24,633 test samples.

For simplicity, these datasets are denoted as CUB, CAR, AIR and NAB, respectively.
All our experiments follow the default training/testing split of these datasets. 
Following the evaluation protocol of the prior self-supervised fine-grained visual categorization work \cite{shu2023learning}, the training set of a fine-grained dataset is used to implement the self-supervised pre-training for the proposed CMD. After that, the pre-trained self-supervised fine-grained representation is fine-tuned in a linear way to support FGVC.

%Top-1 accuracy is the commonly-used metric for FGVC. 
Following \cite{shu2023learning}, the linear probing from the self-supervised representation is evaluated by the top-1 accuracy.
On the other hand, the image retrieval setting is also used for evaluation.
As a kind of nearest neighbor classification, it helps benchmark whether the pre-trained class-agnostic representation learned by SSL is discriminative for fine-grained categories. 
Rank-1, Rank-5 and mAP are reported.
Following \cite{shu2023learning}, 
given the learned features, one image per time is used as the query image to retrieve the rest images
in the test set. This operation goes through all the samples in the test set, and the average is reported.

\subsection{Comparison with Self-supervised Methods}

The proposed method is compared with a variety of self-supervised visual representation learning methods (SimCLR \cite{chen2020simple}, BYOL \cite{grill2020bootstrap}, MoCo v2 \cite{chen2020improved}, DINO \cite{caron2021emerging}, BarlowTwins \cite{zbontar2021barlow}, SimSiam \cite{chen2021exploring}, MAE \cite{he2022masked}, EsViT \cite{li2021efficient}, VICReg \cite{bardes2022vicreg}, I-JEPA \cite{assran2023self}, EPM \cite{feng2023evolved}), a self-supervised fine-grained category discovery method InfoSieve \cite{rastegar2024learn} and a contemporary self-supervised FGVC method LCR \cite{shu2023learning}. 

\subsubsection{On Linear Probing}
Table~\ref{SOTASSL}~reports the top-1 accuracy of the proposed method and existing self-supervised learning methods. 
On both CUB and CAR, it leads to a performance gain of 10.14\% and 11.13\% against the second best-performed method LCR \cite{shu2023learning}, which is a contemporary self-supervised FGVC method.
At least more than 20\% accuracy gain can be observed when compared with other self-supervised visual representation learning methods.
On AIR, it leads to a performance gain of 5.80\% against the second best-performed method EsViT \cite{li2021efficient}. 
Also, it shows a 8.52\% performance gain against LCR \cite{shu2023learning}, and shows at least more than 12\% performance gain against other self-supervised visual learning methods.
Similarly, on NAB, it shows a 25.60\% improvement against the second-best method, and around 40\% improvement against other self-supervised visual representation learning methods.
On the other hand, when using ResNet-50 backbone that has inferior feature representation ability than Swin Transformer, the proposed CMD still significantly outperforms all the compared methods on all the three datasets by up to 5.07\%.
When using ViT-B-16 backbone, the proposed CMD outperforms the other latest methods with the same backbone by up to 9.62\%, 12.67\%, 5.68\% and 25.55\% on CUB, CAR, AIR and NAB, respectively.

\begin{table*}[!t]
%\footnotesize
\begin{center}
    \caption{Comparison with the other state-of-the-art self-supervised learning frameworks with linear probing Top-1 accuracy (in \%), on the CUB, CAR, AIR and NAB datasets. 
    By default, the compared results are directly cited from \cite{shu2023learning}.
    $\dag$: Our implementation by official code and default parameter settings.
    '*': Only one decimal result is reported. '-': Neither official implementation nor source code.
    Parameters (Params., in million) \& throughput (im/s) are both of the feature extractor, and are calculated on a NVIDIA V100 GPU with 128 samples per forward. Results marked in \textbf{bold}, in \textcolor{red}{red} and in \textcolor{blue}{blue} are the best results from all the compared methods, ViT-B-16 backbone and ResNet-50 backbone.} 
\resizebox{0.75\linewidth}{!}{
\begin{tabular}{c|c|c|cc|cccc}
\hline
\multirow{2}{*}{Method} & \multirow{2}{*}{Proc. \& Year} & \multirow{2}{*}{Architectures} & \multirow{2}{*}{Para.} & \multirow{2}{*}{im/s} &  \multicolumn{4}{c}{Linear Top-1 Accuracy} \\
\cline{6-9}
~ & ~ & ~ & ~ & ~ & CUB & CAR & AIR & NAB \\
\hline
\multirow{2}{*}{Supervised} & CVPR2016 & ResNet-50 & 23 & 1237 & 81.34 & 91.02 & 87.13 & 82.23 \\
~ & ICCV2021 & Swin-Tiny$\dag$ & 28 & 808 & 91.09 & 93.59 & 93.43 & 87.46 \\
\hline
DINO \cite{caron2021emerging} & CVPR2021 & ResNet-50 & 23 & 1237 & 16.66 & 10.51 & 12.93 & 10.98$\dag$ \\
Barlow \cite{zbontar2021barlow} & ICML2021 & ResNet-50 & 23 & 1237 & 33.45 & 31.91 & 34.77 & 25.36$\dag$ \\
VICReg \cite{bardes2022vicreg} & ICLR2022 & ResNet-50 & 23 & 1237 & 37.78 & 30.80 & 36.00 & 32.49$\dag$ \\
SimCLR \cite{chen2020simple} & ICML2020 & ResNet-50 & 23 & 1237 & 38.39 & 49.41 & 45.22 & 38.71$\dag$ \\
BYOL \cite{grill2020bootstrap} & NeurIPS2020 & ResNet-50 & 23 & 1237 & 39.27 & 45.21 & 37.62 & 31.83$\dag$ \\
SimSiam \cite{chen2021exploring} & CVPR2021 & ResNet-50 & 23 & 1237 & 39.97 & 58.89 & 43.06 & 33.40$\dag$ \\
MAE \cite{he2022masked} & CVPR2022 & ResNet-50 & 23 & 1237 & 7.68$\dag$ & 12.46$\dag$ & 14.85$\dag$ & 11.56$\dag$ \\
MoCo v2 \cite{chen2020improved} & ArXiv2020 & ResNet-50 & 23 & 1237 & 68.30 & 58.43 & 52.54 & 47.17$\dag$ \\
BYOL+LCR \cite{shu2023learning} & CVPR2023 & ResNet-50 & 23 & 1237 & 51.20 & 50.64 & 45.94 & - \\
EsViT \cite{li2021efficient} & ICLR2022 & Swin-Tiny & 28 & 808 & 61.67$\dag$ & 58.25$\dag$ & 59.59$\dag$ & 43.55$\dag$ \\
LCR \cite{shu2023learning} & CVPR2023 & ResNet-50 & 23 & 1237 & 71.31 & 60.75 & 55.87 & 50.94$\dag$ \\
I-JEPA \cite{assran2023self} & CVPR2023 & ViT-B-16 & 85 & 321 & 54.09$\dag$ & 49.66$\dag$ & 48.83$\dag$ & 39.62$\dag$ \\
EPM \cite{feng2023evolved} & CVPR2023 & ViT-B-16 & 85 & 321 & 57.36$\dag$ & 51.65$\dag$ & 49.61$\dag$ & 41.38$\dag$ \\
InfoSieve \cite{rastegar2024learn} & NeurIPS2024 & ViT-B-16 & 85 & 321 & 69.4* & 55.7* & 56.3* & 48.95$\dag$ \\
\hline
\multirow{3}{*}{CMD (Ours)} & \multirow{3}{*}{2024} & ResNet-50 & 23 & 1237 & \textcolor{blue}{76.38} & \textcolor{blue}{65.49} & \textcolor{blue}{60.21} & \textcolor{blue}{73.13} \\
~ & & ViT-B-16 & 85 & 321 & \textcolor{red}{79.02} & \textcolor{red}{68.37} & \textcolor{red}{61.98} & \textcolor{red}{74.50} \\
~ & & Swin-Tiny & 28 & 808 & \textbf{81.45} & \textbf{71.87} & \textbf{64.39} & \textbf{76.54} \\
\hline
\end{tabular}
}
%\vspace{-.5cm}
\label{SOTASSL}
\end{center} 
\end{table*}

\subsubsection{On Image Retrieval}
Table~\ref{retrivSSL}~reports the image retrieval outcomes of the proposed method, the contemporary LCR \cite{shu2023learning}, two SSL methods (MoCo v2 \cite{chen2020improved}, EsViT \cite{li2021efficient}) and the fully-supervised ResNet-50.
On CUB, the proposed method significantly outperforms all the compared methods, with a 20.19\% gain of Rank-1, 13.43\% gain of Rank-5 and 6.45\% gain of mAP, respectively. 
Similarly, on AIR, it leads to a performance gain of 9.34\% on Rank-1, 6.83\% on Rank-5 and 13.18\% on mAP, respectively.
On Cars, its Rank-1 and Rank-5 are slightly inferior to LCR \cite{shu2023learning}. 
However, its mAP metric achieves 16.21\%, which is 7.34\% higher than LCR \cite{shu2023learning}.
On NAB, it significantly outperforms all the compared methods. Specifically, the improvement over the second-best is 16.95\% on Rank-1, 16.69\% on Rank-5 and 19.27\% on mAP, respectively.

The inferior performance of the proposed CMD than LCR \cite{shu2023learning} may be explained from the less realistic samples in CAR. Specifically, many samples in CAR are synthetic, where the background of a car is artificially set to pure white. As the proposed CMD relies more on the per-instance responses in an image, the artificially low contrast in CAR makes the instance representation less discriminative, leading to a slight performance degradation. In contrast, the experimental outcomes on the NAB dataset further demonstrate the effectiveness of the proposed CMD on large-scale and photo-realistic images, which is more suitable to the real-world FGVC scenarios.

\begin{table}[!t]
\begin{center}
    \caption{Performance comparison between the proposed method and other methods under the image retrieval setting. Evaluation metrics include Rank-1, Rank-5 and mAP (presented in percentage \%). Experiments are conducted on the CUB, CAR, AIR and NAB datasets. 
    }
    \label{retrivSSL}
\resizebox{\linewidth}{!}{
\begin{tabular}{c|c|c|ccc}
\hline
\multirow{2}{*}{Dataset} & \multirow{2}{*}{Method} & \multirow{2}{*}{Architectures} &  \multicolumn{3}{c}{Retrieval} \\
\cline{4-6}
~ & ~ & ~ & Rank-1 & Rank-5 & mAP \\
\hline
\multirow{5}{*}{CUB} & Supervised & ResNet-50 & 10.65 & 29.32 & 5.09 \\
~ & MoCo v2 \cite{chen2020improved} & ResNet-50 & 17.07 & 41.46 & 8.13 \\
~ & EsViT \cite{li2021efficient} & Swin-Tiny & 44.88 & 72.83 & 20.54 \\
~ & LCR \cite{shu2023learning} & ResNet-50 & 49.69 & 75.23 & 24.01 \\
~ & CMD (Ours) & Swin-Tiny & \textbf{69.88} & \textbf{88.66} & \textbf{30.46} \\
%~ & ~ & ~ & \textcolor{red}{+8.52} & \textcolor{red}{+7.34} & \textcolor{red}{+3.19} \\
\hline
\multirow{5}{*}{Cars} & Supervised & ResNet-50 & 4.91 & 16.98 & 2.34 \\
~ & MoCo v2 \cite{chen2020improved} & ResNet-50 & 10.94 & 29.57 & 3.12 \\
~ & EsViT \cite{li2021efficient} & Swin-Tiny & 23.15 & 45.97 & 4.54 \\
~ & LCR \cite{shu2023learning} & ResNet-50 &  \textbf{34.56} & \textbf{60.75} & 8.87 \\
~ & CMD (Ours) & Swin-Tiny & 32.66 & 53.91 & \textbf{16.21} \\
%~ & ~ & ~ & \textcolor{blue}{-1.90} & \textcolor{blue}{-6.84} & \textcolor{red}{+7.34} \\
\hline
\multirow{5}{*}{Aircraft} & Supervised & ResNet-50 & 5.16 & 14.22 & 2.61 \\
~ & MoCo v2 \cite{chen2020improved} & ResNet-50 & 19.38 & 39.90 & 6.30 \\
~ & EsViT \cite{li2021efficient} & Swin-Tiny & 23.15 & 45.97 & 4.54 \\
~ & LCR \cite{shu2023learning} & ResNet-50 & 34.33 & 61.09 & 15.43 \\
~ & CMD (Ours) & Swin-Tiny & \textbf{43.67} & \textbf{67.92} & \textbf{29.61} \\
\hline
\multirow{5}{*}{NAB} & Supervised & ResNet-50 & 5.08 & 17.29 & 2.57 \\
~ & MoCo v2 \cite{chen2020improved} & ResNet-50 & 16.51 & 37.29 & 5.08 \\
~ & EsViT \cite{li2021efficient} & Swin-Tiny & 37.89 & 61.92 & 16.75 \\
~ & LCR \cite{shu2023learning} & ResNet-50 & 46.73 & 68.95 & 25.68 \\
~ & CMD (Ours) & Swin-Tiny & \textbf{63.68} & \textbf{85.64} & \textbf{44.95} \\
\hline
\end{tabular}}
\end{center} 
\end{table}

\subsection{Compared with Other Alternative Solutions}

Some localization based self-supervised learning pipelines may also learn fine-grained representation.
For a more boarder comparison, these recent self-supervised localization pipelines  \cite{shu2023learning,shu2022improving,zhao2021distilling,wu2021align,huang2022learning,peng2022crafting} are also benchmarked on self-supervised FGVC.
Table~\ref{SOTAother}~reports their performance. 
The proposed method significantly outperforms all the compared methods in terms of the top-1 accuracy on the three benchmarks, superior to the second best-performed method by 8.61\%, 8.16\% and 8.31\%, respectively. 
On NAB, it shows a 24.19\% improvement against the second-best method, and around 30\% improvement against other self-supervised visual representation learning methods.

Under the image retrieval setting, it outperforms the second best-performed method by 20.52\% and 8.72\% on CUB and AIR, respectively. On Cars, its Rank-1 metric is slightly inferior to LCR \cite{shu2023learning} and its two variations, but is still significantly higher than other methods by $\textgreater$ 20\%.
On the other hand, when using the same ResNet-50 backbone, it still outperforms all the compared methods on the top-1 accuracy on all three datasets, and the Rank-1 metric on CUB and AIR.
On NAB, it significantly outperforms all the compared methods. Specifically, the improvement over the second-best is 31.76\% on Rank-1.

\begin{table*}[!t]
%\footnotesize
\begin{center}
    \caption{Comparison with the alternatives from self-supervised localization with linear probing Top-1 accuracy (in \%) and Rank-1 retrieval (in \%), on CUB-200-2011, Stanford Cars, FGVC Aircraft and NA-Birds.
    Results marked in \textbf{bold} and in \textcolor{blue}{blue} are the best results from all the compared methods and the best results from ResNet-50 backbone. 
    By default, the compared results are directly cited from \cite{shu2023learning}.
    $\dag$: Our implementation by official code and default parameter settings; '-': Neither report nor with official source code.
    }
    %\vspace{-.2cm}
\resizebox{0.95\linewidth}{!}{
\begin{tabular}{c|c|cccc|cccc}
\hline
\multirow{2}{*}{Method} & \multirow{2}{*}{Architectures} &  \multicolumn{4}{c|}{Linear Top-1 Classification} & \multicolumn{4}{c}{Rank-1 Retrieval} \\
\cline{3-10}
~ & ~ & CUB & CAR & AIR & NAB & CUB & CAR & AIR & NAB \\
\hline
%MoCo v2 \cite{chen2020improved} & ResNet-50 & 68.30 & 58.43 & 52.54 & 17.07 & 10.94 & 19.38 \\
MoCo v2 + Bilinear \cite{shu2023learning} & ResNet-50 & 68.44 & 58.06 & 53.01 & 46.72$\dag$ & 41.27 & 30.89 & 30.80 & 19.89$\dag$ \\
SAM-SSL \cite{shu2022improving} & ResNet-50 & 68.59 & 58.49 & 52.97 & 45.14$\dag$ & 18.38 & 14.26 & 21.72 & 19.08$\dag$\\
SAM-SSL + Bilinear \cite{shu2023learning} & ResNet-50 & 71.56 & 59.12 & 55.12 & 48.63$\dag$ & 44.20 & \textbf{35.38} & 32.10 & 22.36$\dag$ \\
DiLo \cite{zhao2021distilling} & ResNet-50 + FPN & 64.14 & - & - & - & - & - & - & - \\
CVSA \cite{wu2021align} & ResNet-50 + BYOL \cite{grill2020bootstrap} & 65.02 & - & - & - & - & - & - & - \\
LEWEL \cite{huang2022learning} & ResNet-50 & 69.27 & 59.02 & 54.33 & 48.39$\dag$ & 19.23 & 12.01 & 20.67 & 19.57$\dag$\\
ContrastiveCrop \cite{peng2022crafting} & ResNet-50 & 68.82 & 61.66 & 54.40 & 49.16$\dag$ & 18.71 & 13.61 & 20.88 & 20.19$\dag$ \\
%\hline
%LCR \cite{shu2023learning} & ResNet-50 & 71.31 & 60.75 & 55.87 & 49.69 & 34.56 & 34.33 \\
%LCR + MultiTask \cite{shu2023learning} & ResNet-50 & 68.56 & 58.55 & 52.87 & 17.62 & 13.98 & 22.23 \\
LCR + MultiTask \cite{shu2023learning} & ResNet-50 & 72.84 & 63.71 & 56.08 & 52.35$\dag$ & 49.36 & 33.55 & 34.95 & 31.92$\dag$ \\
\hline
\multirow{2}{*}{CMD (Ours)} & ResNet-50 & \textcolor{blue}{76.38} & \textcolor{blue}{65.49} & \textcolor{blue}{60.21} & \textcolor{blue}{73.13} & \textcolor{blue}{60.08} & 31.95 & \textcolor{blue}{39.56} & \textcolor{blue}{59.35} \\
~ & Swin-Tiny & \textbf{81.45} & \textbf{71.87} & \textbf{64.39} & \textbf{76.54} & \textbf{69.88} & 32.66 & \textbf{43.67} & \textbf{63.68} \\
\hline
\end{tabular}
}
\vspace{-.2cm}
\label{SOTAother}
\end{center} 
\end{table*}

\begin{table}[!t]
\begin{center}
\caption{Ablation studies on each component. Metric Top-1 accuracy (in \%) on CUB, CAR and AIR.}
\label{ablation}
\resizebox{\linewidth}{!}{
\begin{tabular}{ccccc|ccc}
\hline
\multicolumn{5}{c|}{Component} & \multicolumn{3}{c}{Linear Top-1 Accuracy} \\
\hline
$\mathcal{L}_{V}$ & $MIL$ & $\mathcal{L}_{R}$ & $\mathcal{L}_{T}$ & $\mathcal{L}_{S}$ & CUB & CAR & AIR \\
\hline
\checkmark & & & & & 43.72 & 46.03 & 50.68 \\
\checkmark & \checkmark & & & & 48.61 & 51.15 & 53.95 \\
\checkmark & \checkmark & \checkmark & & & 61.98 & 60.51 & 59.85 \\
\checkmark & \checkmark & \checkmark & \checkmark & & 65.72 & 62.96 & 61.05 \\
\checkmark & \checkmark & \checkmark & \checkmark & \checkmark & \textbf{81.45} & \textbf{71.87} &  \textbf{64.39}\\
\hline
\end{tabular}}
\end{center} 
\end{table}

\begin{table}[!t]
\begin{center}
\caption{Ablation studies on the label proportions of CMD and other methods on CUB and AIR.}
\label{rateSSL}
%\resizebox{\linewidth}{!}{
\begin{tabular}{c|c|ccc}
\hline
\multirow{2}{*}{Dataset} & \multirow{2}{*}{Method} &  \multicolumn{3}{c}{label proportion} \\
\cline{3-5}
~ & ~ & 100\% & 50\% & 20\% \\
\hline
\multirow{5}{*}{CUB} & ResNet-50 & 68.17 & 58.99 & 46.54 \\
~ & MoCo v2 \cite{chen2020improved} & 68.30 & 60.96 & 46.91 \\
~ & LCR \cite{shu2023learning} & 71.31 & 66.52 & 55.33 \\
~ & CMD (Ours) & \textbf{81.45} & \textbf{76.44} &  \textbf{63.70}\\
~ & ~ & \textcolor{red}{+10.14} & \textcolor{red}{+9.92} & \textcolor{red}{+8.37} \\
\hline
\multirow{5}{*}{AIR} & ResNet-50 & 47.38 & 37.83 & 28.20 \\
~ & MoCo v2 \cite{chen2020improved} & 52.54 & 45.52 & 35.17 \\
~ & LCR \cite{shu2023learning} & 55.87 & 48.22 & 38.55 \\
~ & CMD (Ours) & \textbf{64.39} & \textbf{55.56} & \textbf{41.74} \\
~ & ~ & \textcolor{red}{+8.52} & \textcolor{red}{+7.34} & \textcolor{red}{+3.19} \\
\hline
\end{tabular}
%}
%\vspace{-0.6cm}
\end{center} 
\end{table}

\begin{table}[!t]  
    \centering
    \caption{Impact of $\lambda_1$ on linear probing of the proposed CMD. Top-1 accuracy is presented in ($\%$). Experiments conducted on CUB.}
    \begin{tabular}{c|ccccc} 
    %\toprule[2pt]
    \hline
    $\lambda_1$ & 10 & 1 & 0.1 & 0.01 & 0.001 \\
    \hline
    top-1 acc. & 78.03 & 80.82 & 81.45 & 81.36 & 79.25 \\
    \hline
    %\bottomrule[2pt]
    \end{tabular} 
    \label{ablationeach}
\end{table}

\subsection{Ablation Studies}

\subsubsection{On Each Component}
Table~\ref{ablation}~reports the each component's impact
on the top-1 accuracy for the entire framework.
The baseline is a DINO based framework that only has two image-level crops for both student and teacher net, denoted as $\mathcal{L}_{V}$.
On top of it, the impact of multi-instance learning ($MIL$), region-level crops ($\mathcal{L}_{R}$), inter-level distillation for teacher net ($\mathcal{L}_{T}$), and inter-level distillation for student net ($\mathcal{L}_{S}$), are all studied.
It can be seen that,
the introduce of MIL 
leads to a performance gain of 4.89\%, 5.12\% 
and 3.27\% on CUB, CAR, AIR, respectively.
The introduce of region-level crops can significantly
improve the fine-grained representation, yielding a performance gain of 13.37\% and 9.36\% on CUB and Cars, respectively. It indicates that the joint learning of image-level and region-level crops, although straight-forward, significantly benefits the fine-grained representation.

On the other hand, the inter-level multi-instance distillation on teacher net (denoted as $\mathcal{L}_{T}$) leads to a performance gain of 3.74\%, 2.45\% and 1.20\% on CUB, CAR and AIR, respectively, while inter-level multi-instance distillation on student net (denoted as $\mathcal{L}_{S}$) 
leads to a performance gain of 15.73\%, 8.91\% and 3.34\% on CUB, CAR and AIR. 

\begin{figure}[!t]
  \centering
\includegraphics[width=0.45\textwidth]{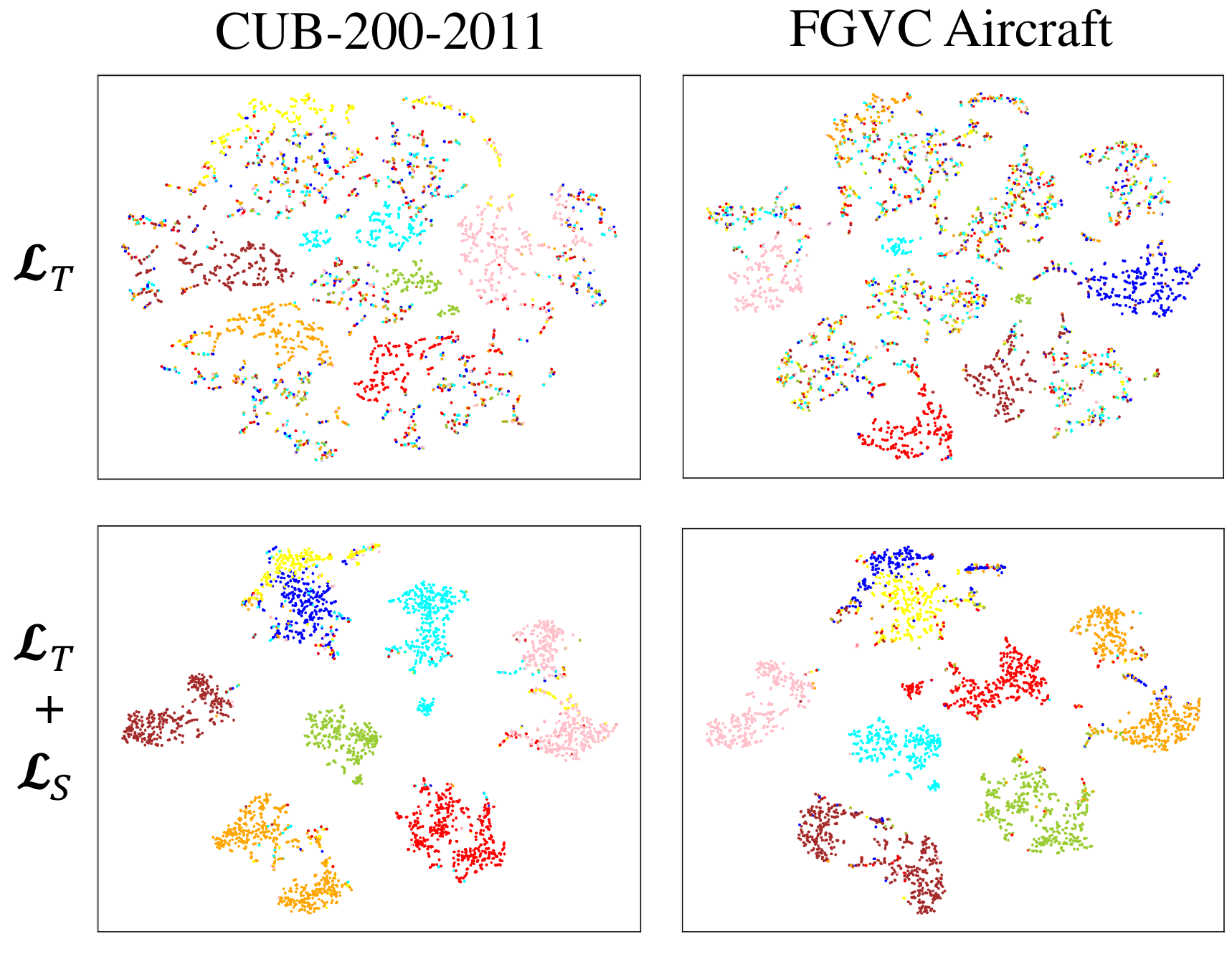} 
%\vspace{-.3cm}
\caption{T-SNE visualization of the features space when only implementing inter-level multi-instance distillation on teacher net ($\mathcal{L}_{T}$) and on both teacher and student net ($\mathcal{L}_{T}$+$\mathcal{L}_{S}$).
}
%\vspace{-.5cm}
\label{tsnevis}
\end{figure}

To better understand the much more significant improvement of $\mathcal{L}_{S}$ than $\mathcal{L}_{T}$, an additional t-SNE visualization for the feature space is shown in Fig.~\ref{tsnevis}. 
Specifically, we randomly select eight fine-grained categories for each dataset, and extract the feature embeddings from $\mathcal{L}_{T}$ and $\mathcal{L}_{T}+\mathcal{L}_{S}$, respectively.
The inter-level multi-instance distillation for the student net ($\mathcal{L}_{S}$) allows the feature embeddings from different fine-grained categories (in different color types) to be more separated, which significantly improves the feature separation of fine-grained categories.

\subsubsection{Intra-level v.s. Inter-level Distillation}

Another important aspect to analyze the contribution of each component is through the proposed Intra-level and Inter-level Multi-instance Distillation, according to the per-component results in Table~\ref{ablation}. 
The intra-level multi-instance distillation, consisting of $\mathcal{L}_{V}$, $MIL$ and $\mathcal{L}_{R}$, leads to a top-1 accuracy improvement of 18.26\%, 14.48\% and 9.17\% on CUB, CAR and AIR, respectively.
In contrast, the inter-level multi-instance distillation, consisting of $\mathcal{L}_{T}$, $MIL$ and $\mathcal{L}_{S}$, leads to a top-1 accuracy improvement of 19.47\%, 11.36\% and 4.54\% on CUB, CAR and AIR, respectively.
Overall, both intra- and inter- level multi-instance distillation lead to a clear performance improvement for self-supervised fine-grained visual categorization.
On AIR dataset, the impact of intra-level multi-instance distillation is slightly higher than inter-level multi-instance distillation (9.17\% v.s. 4.54\%), which may be explained from the less fine-grained categories in AIR than the rest two datasets. 
The less fine-grained categories of a dataset demands a representation to be more discriminative to the image- and region- level crops, instead of the semantic relation between each fine-grained categories.
In fact, the intra-level multi-instance distillation focuses on the relation between image- and region- level crops, which plays a more pivotal role.

\subsubsection{On Region-level Crop Number}
\begin{figure}[!t]
\centering
\includegraphics[width=0.45\textwidth]{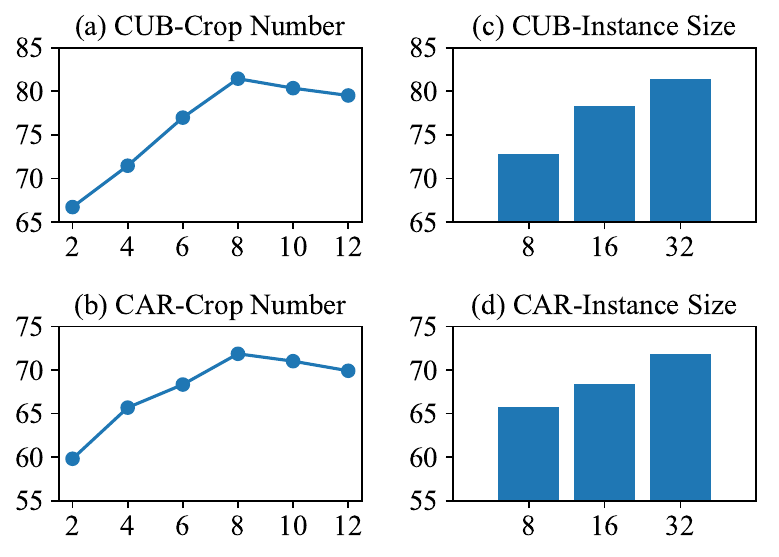} 
%\vspace{-.3cm}
\caption{
Ablation studies on region-level crop number and instance size.
%Ablation studies on region-level crop number ((a) on CUB and (b) on CAR) and instance size ((c) on CUB and (d) on CAR).
}
%\vspace{-.5cm}
\label{ablationvis}
\end{figure}

By default the region-level crop number for both the teacher net and student net is 8.
To extensively investigate the impact of region-level crop number, we test the scenarios when the framework has 2, 4, 6, 8, 10 and 12 region-level crops, and report the results on both CUB and Cars. 
The results are shown in Fig.~\ref{ablationvis}~(a) and (b), respectively. 
When the region-level crop numbers are very few, \eg, 2 and 4, the performance of the proposed CMD degrades greatly, \eg, 66.71\% and 71.46\% on CUB.
When the region-level crop number becomes larger, the performance of the proposed CMD becomes more stable, and achieves its maximum on the size of 8.

\begin{figure*}[!t]
  \centering
   \includegraphics[width=1.0\linewidth]{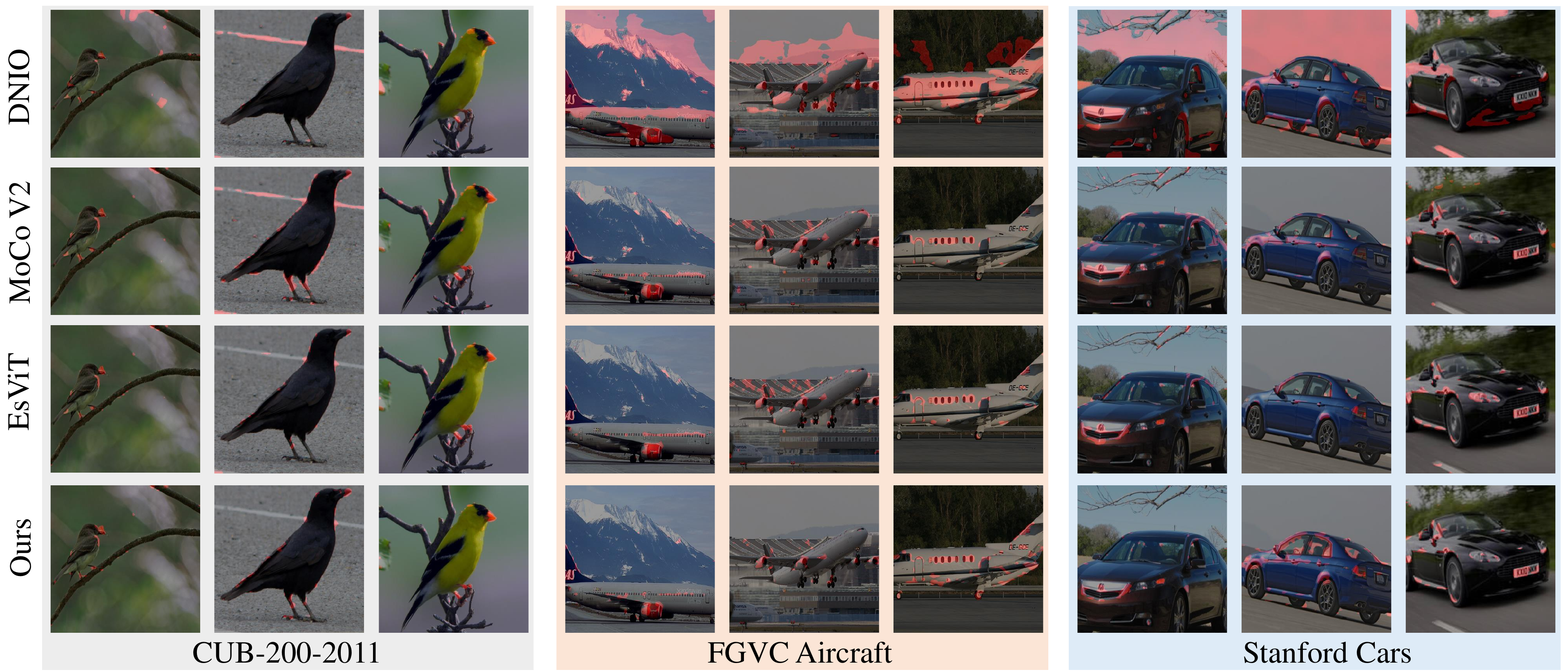}
   \vspace{-0.5cm}
   \caption{
   Activation comparison of the Fine-grained patterns between the proposed CMD and existing SSL methods. Following existing FGVC \cite{zhang2019learning,touvron2019fixing,ji2020attention,du2020fine,zhao2021part}, the last-layer feature from the backbone is visualized by GradCAM.
   }
   \label{heatmapvis}
   %\vspace{-0.2cm}
\end{figure*}

\begin{figure*}[!t]
  \centering
   \includegraphics[width=1.0\linewidth]{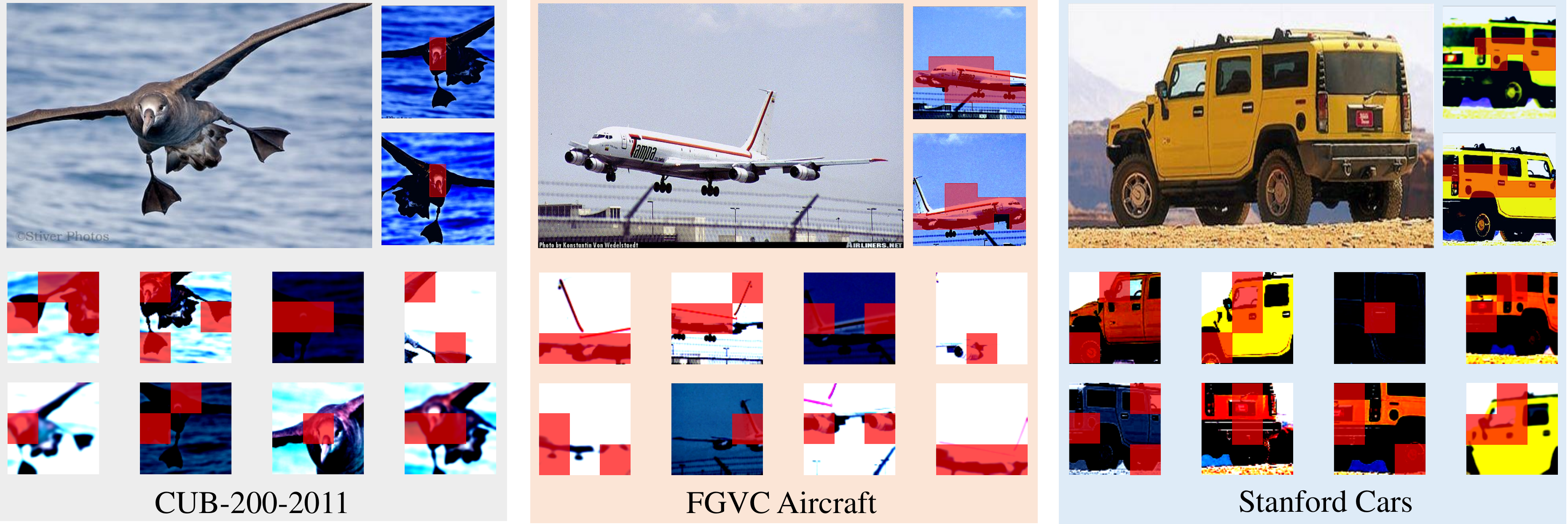}
   \vspace{-0.4cm}
   \caption{By the formulation of multiple instance learning, the key instances (in red rectangular) in both region-level and image-level crops are properly activated.
   }
   \label{insvis}
   %\vspace{-0.5cm}
\end{figure*}

\subsubsection{On Instance Sizes}
By default the instance size is set $32\times32\times3$ for the framework. To investigate 
the robustness of the multi-instance modeling, we systematically test the situation when the instance size is 2, 4, 8, 16 and 32, respectively.
Note that the instance size has to be the common divisor of 96 and 224, so that 32 is the max common divisor that the size can reach.
The results on CUB and CAR are shown in Fig.~\ref{ablationvis}. When the instance size is 2 or 4, the CMD collapses and does not yield meaningful results.
Also, when the instance size is 8, the performance declines to 72.75\% and 65.49\% on CUB and Cars.
These observations can be explained from \textit{Curse of Dimensionality}, as the too-small instance size significantly increases the dimension of feature embedding.
In contrast, when the instance size is larger, \eg, 16 or 32, the performance of CMD becomes stable owing to the proper size of feature embedding.

\begin{figure}[!t]
  \centering
   \includegraphics[width=1.0\linewidth]{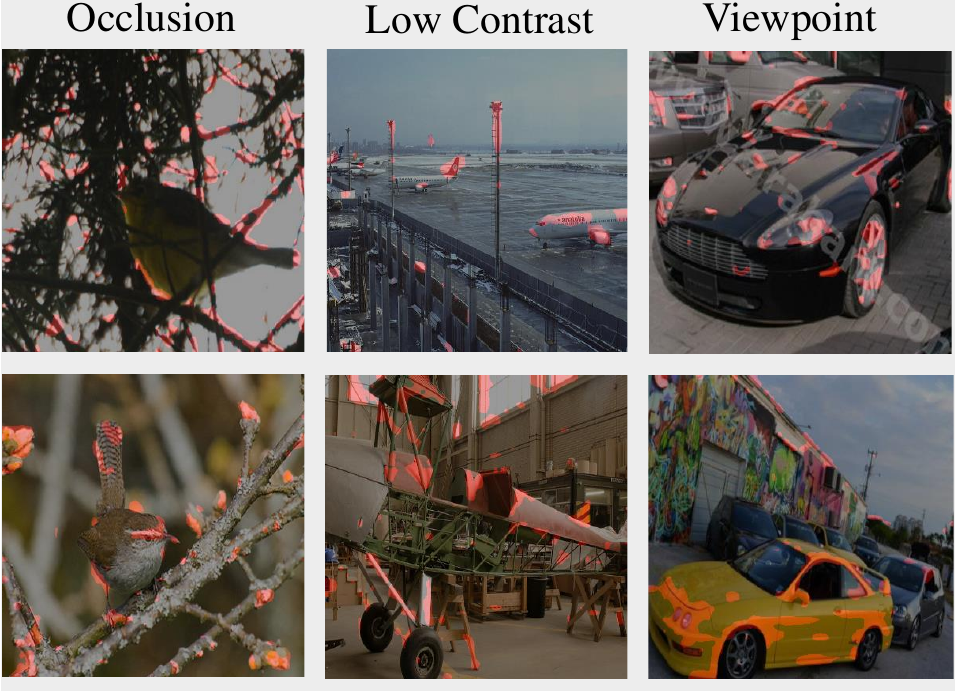}
   \vspace{-0.6cm}
   \caption{
   Some failure cases, caused by occlusion, low contrast and varied viewpoints.
   }
   \label{fail}
   %\vspace{-0.2cm}
\end{figure}

\subsubsection{On Different Label Proportions}
When training the linear classifier, the number of training samples significantly impacts the performance. 
Table~\ref{rateSSL}~reports the performance of the proposed method and the contemporary LCR \cite{shu2023learning}, a self-supervised visual representation method MoCo v2 \cite{chen2020improved} and the fully-supervised ResNet-50.
On CUB, when the label proportion is 100\%, 50\% and 20\%, the proposed method shows a performance gain of 10.14\%, 9.92\% and 8.37\% against the second best-performed method. 
On AIR, the performance gain against the second best-performed method is 8.52\%, 7.34\% and 3.19\% under the proportion of 100\%, 50\% and 20\%.

\subsubsection{On Each Data Augmentation Operation}
Modern SSL pipelines usually rely on multiple types of data augmentation to generate the crop pairs, so does the proposed CMD. Keeping the same as the prior works \cite{li2021efficient}, five commonly-used data augmentation operations, namely, Random \& Horizontal Flipping, Color Jittering, Random-resized Cropping, Gaussian Blurring and Solarization, are used in the proposed CMD. 
For simplicity, they are denoted as RF, CJ, RC, GB and SZ, respectively.
An ablation study is conducted to analyze the impact of each operation. The results are reported in Table~\ref{ablationaug}. In general, each of the five data augmentation operations positively contributes to the performance on self-supervised FGVC, yielding a total top-1 accuracy improvement of 3.51\%, 3.31\% and 3.12\% on CUB, CAR and AIR, respectively. 
Specifically, the individual improvement of RF, CJ, RC, GB and SZ on CUB is 0.58\%, 0.78\%, 0.73\%, 0.82\% and 0.60\%,  respectively, in top-1 accuracy.
The individual improvement of RF, CJ, RC, GB and SZ on CAR is 0.54\%, 0.82\%, 0.67\%, 0.71\% and 0.57\%, respectively, in top-1 accuracy.
The individual improvement of RF, CJ, RC, GB and SZ on AIR is 0.51\%, 0.79\%, 0.84\%, 0.57\% and 0.41\%, respectively.

\begin{table}[!t]
\begin{center}
\caption{Ablation studies on each data augmentation operation. Metric Top-1 accuracy (in \%) on CUB, CAR and AIR. }
\label{ablationaug}
\resizebox{\linewidth}{!}{
\begin{tabular}{ccccc|ccc}
\hline
\multicolumn{5}{c|}{Operation} & \multicolumn{3}{c}{Linear Top-1 Accuracy} \\
\hline
$RF$ & $CJ$ & $RC$ & $GB$ & $SZ$ & CUB & CAR & AIR \\
\hline
 & & & & & 77.94 & 68.56 & 61.27 \\
\checkmark & & & & & 78.52 & 69.10 & 61.78 \\
\checkmark & \checkmark & & & & 79.30 & 69.92 & 62.57 \\
\checkmark & \checkmark & \checkmark & & & 80.03 & 70.59 & 63.41 \\
\checkmark & \checkmark & \checkmark & \checkmark & & 80.85 & 71.30 & 63.98 \\
\checkmark & \checkmark & \checkmark & \checkmark & \checkmark & \textbf{81.45} & \textbf{71.87} &  \textbf{64.39}\\
\hline
\end{tabular}}
\end{center} 
\end{table}

\subsection{Sensitivity Analysis of $\lambda_1$}

The learning objective of the proposed CMD is dependent on the $\lambda_1$ value in Eq.~\ref{totalloss}.
As reported in Table.~\ref{ablationeach},
we test the settings when $\lambda_1$ takes value from $\{10, 1, 0.1, 0.01 , 0.001\}$ on the CUB-200-2011 dataset, and observe a top-1 accuracy (in percentage \%) of $\{78.03, 80.82, 81.45, 81.36, 79.25\}$.

The results indicate that, a too small $\lambda_1$ can reduce the impact of Inter-level Multi-instance Distillation ($\mathcal{L}_S$ and $\mathcal{L}_T$), while a too large $\lambda_1$ can overwhelm the impact of Intra-level Multi-instance Distillation ($\mathcal{L}_I$ and $\mathcal{L}_R$).
Therefore, we choose a trade-off value 0.1 for $\lambda_1$. 

\subsection{Generalization Ability Test}

\subsubsection{On Different Backbones}
Apart from the use of Swin-Tiny and ResNet-50 backbone in the main submission, 
we further test the generalization
ability of the proposed CMD on other
CNN and ViT backbones, including ResNet-101 \cite{he2016deep}, Convolutional vision Transformer (CvT) \cite{wu2021cvt}, Vision Longformer (ViL) \cite{zhang2021multi} and Swin-Small \cite{liu2021swin}.

The experiments are conducted on FGVC Aircraft dataset, and the results are shown in Tab.~\ref{backbonevis}.
The proposed CMD, after linear probing, achieves a top-1 accuracy of 60.21\%, 61.58\%, 58.71\%, 62.18\%, 64.39\% and 66.25\% on the backbone of ResNet-50, ResNet-101, CvT, ViL, Swin-Tiny and Swin-Small, respectively.
Generally, CMD shows a good generalization on a variety of CNN and ViT based backbones.

\begin{table}[!t]  
    \centering
    \caption{Generalization of the proposed CMD on different backbones. Top-1 accuracy is presented in ($\%$). Experiments conducted on AIR.}
    \begin{tabular}{c|cccccc} 
    %\toprule[2pt]
    \hline
    ~ & Res-50 & Res-101 & CvT & ViL & Swin-T & Swin-S \\
    \hline
    acc. & 60.21 & 61.58 & 58.71 & 62.18 & 64.39 & 66.25 \\
    \hline
    %\bottomrule[2pt]
    \end{tabular} 
    \label{backbonevis}
\end{table}

\begin{table}[!t]
\begin{center}
\caption{Generalization ability on other types of self-supervised learning pipelines. $\mathcal{U}$, $\mathcal{D}$, and $\mathcal{C}$: cluster, knowledge distillation and contrastive learning based paradigm; Metric Top-1 accuracy (in \%) on CUB, CAR and AIR.}
\label{generalization}
\resizebox{1.0\linewidth}{!}{
\begin{tabular}{c|c|c|ccc}
\hline
\multirow{2}{*}{Method} & \multirow{2}{*}{Category} & \multirow{2}{*}{Backbone} & \multicolumn{3}{c}{Linear Top-1 Accuracy} \\
\cline{4-6}
~ & ~ & ~ & CUB & CAR & AIR \\
\hline
Suave \cite{fini2023semi} & $\mathcal{U}$ & \multirow{2}{*}{ResNet-50} & 62.89 & 54.25 & 47.60 \\
Suave + CM (Ours) & $\mathcal{U}$ & ~ & 68.01 & 59.83 & 51.36 \\
\hline
Baseline ($\mathcal{L}_V$) & $\mathcal{D}$ & \multirow{4}{*}{Swin-T} & 43.72 & 46.03 & 50.68 \\
$\mathcal{L}_V$ + $MIL$ & $\mathcal{D}$ & & 48.61 & 51.15 & 53.95 \\
CM (Ours) + $\mathcal{L}_{CL}$ & $\mathcal{C}$ & & 75.38 & 65.87 & 59.62 \\
CMD (Ours) & $\mathcal{D}$ & & \textbf{81.45} & \textbf{71.87} &  \textbf{64.39}\\
\hline
\end{tabular}
}
\end{center} 
\end{table}

\subsubsection{On Other Self-Supervised Learning Paradigm}

We further investigate the generalization of the proposed CMD on other self-supervised learning paradigms, namely, clustering ($\mathcal{U}$) and contrastive learning ($\mathcal{C}$).
Two experiments are designed. The first experiment is under the cluster based paradigm. Suave \cite{fini2023semi} is used as the baseline, and we embed the cross-level multiple instance formulation (denoted as CM) before implementing clustering.
The second experiment is under the contrastive learning paradigm. We keep the all the rest components in the CMD as the same, and only replace all the distillation losses by the contrastive losses (denoted as CM+$\mathcal{L}_{CL}$). The scenario that does not involve the cross-level losses (denoted as $\mathcal{L}_V+MIL$) is also involved for comparison.

Table~\ref{generalization} reports the results. Under the clustering paradigm, the propose cross-level multi-instance formulation leads to a top-1 accuracy improvement of 5.12\%, 5.58\% and 4.76\% on CUB, CAR and AIR, respectively. Besides, the introduction of contrastive learning paradigm over the cross-level multiple instance formulation can also significantly improve the representation ability. Compared with the vanilla self-supervised knowledge distillation pipeline with MIL ($\mathcal{L}_V+MIL$), it leads to a top-1 accuracy improvement of 26.77\%, 14.72\% and 4.77\%, respectively.

\subsubsection{Knowledge Distillation v.s. Contrastive Learning} The replacement of distillation with contrastive learning leads to a top-1 accuracy decline by 6.07\%, 6.00\% and 4.77\% on CUB, CAR and AIR, respectively. These outcomes indicate that the contrastive learning paradigm is less effective than knowledge distillation for self-supervised FGVC.

The explanation may be two-fold. Firstly, self-supervised knowledge distillation does not rely on the concept of negative pairs or samples.
In the context of FGVC, the differences between fine-grained categories are usually subtle, which lead to the confusion between positive and negative samples (e.g, in contrastive learning), 
and therefore degrade performance.Secondly, self-supervised knowledge distillation can learn more discriminative semantic information through the teacher-student network.
The teacher network produces soft representations to guide the student network, which allows the model to further focus on more discriminative and detailed features and to better distinguish the subtle differences between fine-grained categories.
In contrast, contrastive learning only separates the overall representations but does not focus on fine-grained differences. 

\subsection{Effectiveness over Fully-supervised FGVC Methods}

It is beneficial to investigate if the self-supervised FGVC methods, especially the proposed CMD, can improve the representation over existing fully-supervised FGVC methods. However, the fully-supervised FGVC methods usually consist of more modules and trainable parameters than the self-supervised methods, which solely use the image encoder along with a linear classifier.
More modules and trainable parameters can fit the fine-grained category representation much better than the image encoder, which poses a predominant and unfair advantage over the self-supervised FGVC methods. Besides, for self-supervised FGVC methods, they are under a two-stage training pipeline, where the supervision of the category label is only involved in the second stage (\ie, fine-tuning the image encoder with only a linear classifier).
In contrast, fully-supervised FGVC methods are directly trained in a one-stage paradigm with the supervision of the category label.

Therefore, we analyze and compare their effectiveness under the one-stage fully-supervised paradigm.
The student net branch serves as the auxiliary branch in \cite{yang2022fine}. For the embedding from each crop, either region- or local- level, it was processed by a global maximum pooling followed by a multi-layer perceptron (MLP), and then supervised by the category label. These modifications keeps the same as \cite{yang2022fine} for fair evaluation. More implementation and architecture details can refer to \cite{yang2022fine}.
The results are shown in Table~\ref{encoderfull}.
With more trainable parameters, modules and direct supervision, the representation learned by the proposed CMD can also benefit FGVC under the full supervision setting. 
It achieves a top-1 accuracy of 92.0\%, 95.6\% and 94.1\% on CUB, CAR and AIR, respectively. 
It significantly outperforms existing fully-supervised FGVC methods with the same ResNet-50 backbone.
Besides, it achieves a competitive performance with more recent methods (\eg, SIM-OFE~\cite{Sun2024SIM}), where both the methodology design and the ViT-B-16 backbone are more powerful and effective.
These outcomes indicate the scalability and effectiveness of the proposed CMD to full supervision.

\begin{table}[!t]
\begin{center}
\caption{Comparison under the one-stage full supervision paradigm.
The proposed CMD is attached with the global maximum pooling and multi-head MLP. Its teacher and student net is directly supervised by the category label as \cite{yang2022fine} does.
Following the convention, results are directly cited from prior references, and reported in one decimal. '-': not reported.
The best and second-best performance is highlighted in \textbf{bold} and in \underline{underline}, respectively.
Metric Top-1 accuracy (in \%) on CUB, CAR and AIR.}
\label{encoderfull}
\resizebox{1.0\linewidth}{!}{
\begin{tabular}{c|c|ccc}
\hline
\multirow{2}{*}{Method} & \multirow{2}{*}{Backbone} & \multicolumn{3}{c}{Linear Top-1 Accuracy} \\
\cline{3-5}
~ & ~ & CUB & CAR & AIR \\
\hline
%ImageNet & ResNet-50 & 81.3 & 91.0 & 87.1 \\
P-CNN~\cite{han2022pcnn} & VGG-19 & 87.3 & 93.3 & 90.6 \\
DCAL~\cite{zhu2022dual} & ViT-B-16 & \underline{92.0} & 94.7 & 93.3 \\
P2P-Net \cite{yang2022fine} & ResNet-50 & 90.2 & 95.4 & 94.2 \\
ABC-Norm~\cite{Hsu2023ABC} & ResNet-50 & 87.8 & - & 93.2 \\
GCP~\cite{Song2023GCP} & ResNet-50 & 92.5 & 92.5 & 91.0 \\
PMRC~\cite{tang2023weakly} & DenseNet-101 & 91.5 & 95.2 & 94.0 \\
GTF~\cite{zhu2023learning} & DenseNet-161 & 91.5 & - & - \\
MP-FGVC~\cite{jiang2024delving} & ViT-B-16 & 91.8 & - & - \\
SIM-OFE~\cite{Sun2024SIM} & ViT-B-16 & \textbf{92.3} & - & - \\
\hline
\textbf{CMD} + Full Supervision & ResNet-50 & \underline{92.0} & \textbf{95.6} &  \textbf{94.1}\\
\hline
\end{tabular}
}
\end{center} 
\end{table}

\subsection{Visualization}

\subsubsection{On Fine-grained Patterns}
Fig.~\ref{heatmapvis}~provides some visualization of the fine-grained patterns on CUB, AIR and CAR. 
The activation pattern of the proposed CMD is compared with three strong self-supervised learning methods, namely, DINO \cite{caron2021emerging}, MoCo v2 \cite{chen2020improved} and EsViT \cite{li2021efficient}. 
The proposed CMD is more effective to highlight the key local patterns that discern each fine-grained category from the others.
Notably, the quality of the activation pattern can be quite close to existing fully-supervised FGVC methods \cite{zhang2019learning,touvron2019fixing,ji2020attention,du2020fine,zhao2021part}. 

\subsubsection{On Activated Instances}
Fig.~\ref{insvis}~shows the activated \textit{instances} by the proposed CMD on both image-level and region-level crops, both of which are formulated as \textit{bag}.
These instances are located on the regions of the objects that help discriminate them from other fine-grained categories.

\subsubsection{On Failure Cases}
Fig.~\ref{fail}~shows some failure cases of the proposed CMD, where the fine-grained patterns are not properly activated. It is observed that in such cases, the object is captured in conditions such as low contrast, varied viewpoint and occlusion. Nevertheless, such cases are common challenges to learn a robust representation for most vision tasks, not just for self-supervised FGVC or the proposed CMD.

\section{Conclusion}
\label{sec6}

Self-supervised FGVC is an emerging topic.
In this paper, we propose a simple, straight-forward but effective method \textbf{C}ross-level \textbf{M}ulti-instance \textbf{D}istillation (CMD).
It uses the multiple instance formulation to model the relation between image patches and fine-grained semantics.
Both intra-level and inter-level multi-instance distillation are proposed to learn the informative patches that relate to the fine-grained semantics.
Extensive experiments on three publicly available FGVC benchmarks show it significantly outperforms the state-of-the-art SSL and self-supervised FGVC methods.

The proposed CMD has the potential for large-scale pre-training on huge amount of unlabeled fine-grained images,
but also has some limitation. Firstly, the proposed CMD is based on the assumption that the fine-grained patterns rest in several key instances. In some extreme noisy or adverse conditions when the key instance is corrupted, the self-supervised representation learned by CMD can be degraded. Secondly, the relation between bag and instance is computed by mean pooling based aggregation function, which does not consider the individual contribution of each instance.

Our future work is three-fold. First, we will improve the robustness on more challenging scenarios, where the key instances may be degraded by adverse conditions such as noise, low-light and conclusion. Secondly, we will improve the learning scheme so as to weight the contribution of each instance in determining the fine-grained bag representation. Thirdly, we will investigate the generalization ability to other semi- and weakly- supervised learning pipelines.

\section*{Acknowledgment}
This work was supported by the National Natural Science Foundation of China under contracts No.U22B2011 and No.62325111.
The authors would like to thank the editors and the anonymous reviewers, whose insightful suggestions and comments significantly improved our paper.

%\appendix

%The pseudo code of the proposed CMD is given in Algorithm~\ref{proposedalgo}.
%The four learning objective functions for both inter-level and intra-level multi-instance distillation are easy to be implemented and integrated into the existing Self Distillation with No Labels (DINO) pipelines, \eg, DINO\cite{caron2021emerging}, EsViT \cite{li2021efficient}. 

% ref
\bibliographystyle{IEEEtran} %ieeetr国际电气电子工程师协会期刊
\bibliography{ref}

\end{document}